%% file: iclr2025_conference.tex
\documentclass{article} % For LaTeX2e
\usepackage{iclr2025_conference,times}

\input{math_commands.tex}

\definecolor{Gray}{gray}{0.9}
\definecolor{LightCyan}{rgb}{0.88,1,1}
\definecolor{red}{RGB}{144,0,32}
\definecolor{blue}{rgb}{0.06, 0.3, 0.57}
\definecolor{warmblack}{rgb}{0.0, 0.26, 0.26}
\definecolor{purple}{rgb}{0.4, 0.01, 0.24}
\definecolor{tawny}{rgb}{0.8, 0.34, 0.0}
\definecolor{amber}{rgb}{1.0, 0.49, 0.0}
\definecolor{antiquefuchsia}{rgb}{0.57, 0.36, 0.51}
\usepackage{hyperref}
\usepackage{url}
\hypersetup{
    colorlinks=true,
    citecolor=antiquefuchsia,
    linkcolor=warmblack,
    urlcolor=purple,
    }

\usepackage[utf8]{inputenc} % allow utf-8 input
\usepackage[T1]{fontenc}    % use 8-bit T1 fonts
\usepackage{hyperref}       % hyperlinks
\usepackage{url}            % simple URL typesetting
\usepackage{booktabs}       % professional-quality tables
\usepackage{amsfonts}       % blackboard math symbols
\usepackage{nicefrac}       % compact symbols for 1/2, etc.
\usepackage{microtype}      % microtypography
\usepackage{xcolor}         % colors

\usepackage{graphicx}
\usepackage{amsmath}
\usepackage{caption}
\usepackage{subcaption}
\usepackage{algorithm}
 \usepackage{algorithmic}
 \usepackage{listings}
 \usepackage{multirow} 
\usepackage{wrapfig}
\title{Diffusion-based Neural Network Weights\\ Generation}

\usepackage{array}
\usepackage{enumitem}

\usepackage{tabularx}

\author{
    \hspace{-4pt} Soro Bedionita$^{1*}$ \ \ \ Bruno Andreis$^{1}$\thanks{Equal Contribution. Correspondence to: Soro Bedionita <sorobedio@kaist.ac.kr> Bruno Andreis <andries@kaist.ac.kr>, Hayeon Lee <yeonhi926@gmail.com>, Wonyong Jeong <wyjeong@kaist.ac.kr>, Song Chong <songchong@kaist.ac.kr>, Frank Hutter <fh@cs.unifreiburg.de>, Sung Ju Hwang <sjhwang82@kaist.ac.kr>} \ \ \ Hayeon Lee$^{4}$  \ \ \ \textbf{Wonyong Jeong}$^3$ \ \ \ \textbf{Song Chong}$^1$ \\
    \textbf{Frank Hutter}$^2$ \ \ \ \textbf{Sung Ju Hwang}$^{1, 3}$ \\
    $^1$KAIST \ \ $^2$University of Freiburg \ \ $^3$DeepAuto, \ \ $^4$Meta AI 
}

\newcommand{\ourmethod}{\text{D2NWG}\,}

\iclrfinalcopy % Uncomment for camera-ready version, but NOT for submission.
\begin{document}

\maketitle

\begin{abstract}
    \input{sections/abstract}
\end{abstract}

\input{sections/introduction}

\input{sections/related}

\input{sections/approach}

\input{sections/experiments}

\section{Conclusion}
\input{sections/discussion}

% \newpage
% % \clearpage
% {
% \small
% \bibliography{Styles/reference}
% \bibliographystyle{abbrvnat}
% }
% \newpage
% \newpage 
% \section*{NeurIPS Paper Checklist}
% \input{Styles/sections/check_List}

% \newpage
% \appendix
% \section{Appendix / supplemental material}
% \input{Styles/sections/appendix}


\bibliography{iclr2025_conference}
\bibliographystyle{iclr2025_conference}

\input{sections/check_List}
\newpage
\appendix
% \section{Appendix}
\input{sections/appendix}

\end{document}

%% file: math_commands.tex
\usepackage{amsmath,amsfonts,bm}

\def\eqref#1{equation~\ref{#1}}
\def\1{\bm{1}}

\DeclareMathAlphabet{\mathsfit}{\encodingdefault}{\sfdefault}{m}{sl}
\SetMathAlphabet{\mathsfit}{bold}{\encodingdefault}{\sfdefault}{bx}{n}

%% file: sections/abstract.tex
% \begin{abstract}

Transfer learning has gained significant attention in recent deep learning research due to its ability to accelerate convergence and enhance performance on new tasks. However, its success is often contingent on the similarity between source and target data, and training on numerous datasets can be costly, leading to blind selection of pretrained models with limited insight into their effectiveness. To address these challenges, we introduce D2NWG, a diffusion-based neural network weights generation technique that efficiently produces high-performing weights for transfer learning, conditioned on the target dataset. Our method extends generative hyper-representation learning to recast the latent diffusion paradigm for neural network weights generation, learning the weight distributions of models pretrained on various datasets. This allows for automatic generation of weights that generalize well across both seen and unseen tasks, outperforming state-of-the-art meta-learning methods and pretrained models. 
Moreover, our approach is scalable to large architectures such as large language models (LLMs), overcoming the limitations of current parameter generation techniques that rely on task-specific model collections or access to original training data. By modeling the parameter distribution of LLMs, D2NWG enables task-specific parameter generation without requiring additional fine-tuning or large collections of model variants. Extensive experiments show that our method consistently enhances the performance of diverse base models, regardless of their size or complexity, positioning it as a robust solution for scalable transfer learning.

%% file: sections/introduction.tex
\section{Introduction}
\label{sec:intro}
\par In recent years, generative AI models have transformed artificial intelligence, with impactful applications in natural language processing (NLP), audio generation, image creation, and video synthesis~\citep{gozalobrizuela2023surveygenerativeaiapplications}. Among these, diffusion-based models have emerged as a leading approach for generating real-valued data through a denoising process~\citep{ho2020denoising, 9878449, 10377858, 10378581}. These models have achieved remarkable results across diverse applications, including task-conditioned signal generation and image and video synthesis, showcasing their versatility and effectiveness in various domains ~\citep{yang2024diffusionmodelscomprehensivesurvey}. 

\par Extending Denoising Diffusion Probabilistic Models (DDPM) to neural network weight generation will offer a powerful solution for improving transfer learning and model adaptation. By dynamically generating task-specific weights, such an approach enables more efficient training, better generalization, and faster adaptation, unlocking new potential for automatic machine learning (AutoML)~\citep{hutter2019automated, 9579526}. 

\par Generative hyper-representation learning~\citep{schurholtHyperRepresentationsGenerativeModels2022}, which generates neural network weights from pretrained distributions, is gaining traction. While recent methods like latent diffusion-based weight generation~\citep{wang2024neural, schuerholt2024sane} show promise, they are limited to small architectures or arbitrary parameter subsets without clear justification. Furthermore, they focus solely on in-distribution sampling, overlooking the critical need for task- and dataset-conditioned weight generation for unseen tasks. 

\par To address multi-task learning, \citet{nava2023metalearning} and \citet{Zhang_2024} proposed meta-learning methods for weight generation in visual and few-shot tasks. However, these approaches produce suboptimal parameters that still require fine-tuning for both in-distribution and out-of-distribution tasks, limiting their efficiency and adaptability. 
\par Addressing these limitations could unlock more robust, scalable solutions for adaptive learning across diverse tasks. Additionally, most of these methods are focused on vision tasks, particularly image classification, overlooking their potential in large language models (LLMs). Expanding these approaches to LLMs could unlock significant advancements in task-specific weight generation and model adaptation as well as pretrained model optimal parameters space exploration.

\par In this work, we introduce a generative hyper-representation approach and reffered to as \ourmethod,  that leverages latent diffusion to harness knowledge from a vast collection of pretrained neural network weights across diverse datasets. Our method generates dataset- or task-conditioned parameters tailored to specific tasks, serving as a generative retriever for in-distribution tasks without fine-tuning and as a task-adaptive parameter generator for unseen tasks. 

\par We extend this approach to large language models (LLMs) by using diffusion-based weight generation to explore optimal weight spaces. This preserves the integrity of pretrained weights while generating task-specific parameters. We hypothesize that both original and task-specific fine-tuned weights lie within the same manifold, enabling our method to produce diverse, optimized weights without the need for fine-tuning. An overview of our method is depicted in Figure~\ref{fig:overview}.
\input{figures/concept}

Our contributions are as follows:
\begin{itemize}
    \item We introduce a novel neural network weight generation framework using latent diffusion models, generating task-adaptive weights conditioned on datasets or task descriptions.
    \item D2NWG outperforms meta-learning approaches on unseen datasets by generating superior weights.
    \item Our method surpasses the performance of pretrained weights and enables rapid adaptation for both classifier heads and full models.
    \item It scales to small and large datasets, generating weights for architectures with over 400 million parameters including GPT2-Small.
    \item We demonstrate its effectiveness in improving LLM performance by generating task-specific weights from a single pretrained model and our sampled weights based LLAMA3-.1-8B and LLAMA3-.2-1B models ranked among top 2 performing models on the open lm-leaderboard\footnote{\url{https://huggingface.co/spaces/open-llm-leaderboard/open_llm_leaderboard}}
\end{itemize}

%% file: figures/concept.tex
\begin{figure}[t]
\begin{center}
\includegraphics[width=1.0\textwidth]{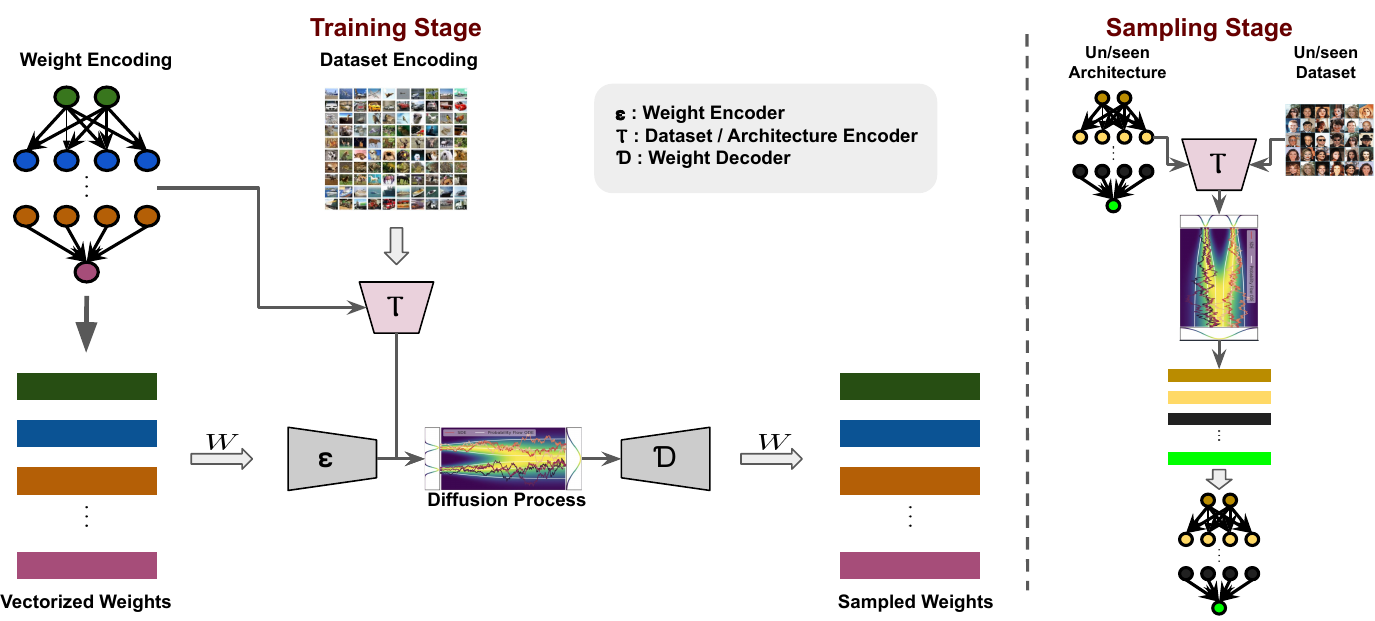}
\end{center}
\vspace{-0.05in}
\caption{\textbf{Stage 1:} VAE Encoder and Decoder training process. \textbf{Stage 2}:  dataset encoder training stage . \textbf{Stage 3:} Dataset conditioned diffusion process. }
%Each stage can be trained independently}
\label{fig:overview}
\end{figure}

%% file: sections/related.tex
\section{Related work}
As neural networks find broader applications, leveraging pretrained weights has become essential for accelerating transfer learning. In this context, generating network weights directly from model zoos, without relying on traditional training pipelines, offers a promising path to improve model efficiency. Several hypernetwork approaches for weight prediction have been proposed to address this challenge~\citep{Chauhan2023ABR, ratzlaff2020hypergan, NIPS2013_7fec306d, ha2016hypernetworks}.

\par{\textbf{Neural Network Parameters Prediction}}:
Recently, \citet{zhang2018graph} introduced Graph Hypernetwork (GHN) to generate weights using a model's directed graph representation. This was enhanced by \citet{knyazev2021parameter} with GHN2, which focused on generating weights across architectures for the same datasets. Similarly, \citet{pmlr-v162-zhmoginov22a} treated weight generation as an autoregressive process, using a transformer to generate weights layer by layer, though this approach is less scalable due to the need for a transformer per layer. Building on this, \citet{knyazev2023canwescale} combined transformer-based techniques with GHN2 to create GHN3, improving generalization across architectures and datasets. However, GHN3’s deterministic sampling limits its generalization, relying heavily on the diversity of training architectures and the pretrained dataset..

\par \textbf{Meta Pretrained Weight Generators}:~\citet{nava2023metalearning} proposed HyperLDM, a generative model for weight generation in visual question answering (VQA) tasks. This model leverages the distribution of weights pretrained in a meta-learning setting and uses latent diffusion for sampling. Similarly, \citet{Zhang_2024} integrated diffusion-based meta-weight generation to enhance adaptation for few-shot learning. While generating pretrained weights through meta-training shows promising results, the meta-learning process can be computationally expensive. Additionally, the meta-pretrained weights are not optimal even for in-distribution evaluation they always require some optimization steps.

\par \textbf{AutoEncoder-based Weight Generators}: \citet{schurholt_self-supervised_2021} proposed learning the distribution of weights by reconstructing them using autoencoder-style architectures. In a follow-up work, \citet{schurholtHyperRepresentationsGenerativeModels2022} introduced a method for learning the distribution of pretrained weights, allowing for unconditional sampling of diverse weights through kernel density estimation. A related approach by \citet{peebles2022learning} involves conditioning weight generation on the target loss using a diffusion transformer framework.

\par \textbf{Diffusion Models}: Denoising Diffusion Probabilistic Models (DDPM)~\citep{NEURIPS2020_4c5bcfec, Rombach2021HighResolutionIS, NEURIPS2022_62868cc2, Croitoru2022DiffusionMI} enable mapping data representations to a Gaussian distribution and vice versa achieving state-of-the art performance in generative modeling. Latent diffusion models~\citep{dhariwal2021diffusion, pmlr-v139-nichol21a, ho2021classifierfree, 10203150, Rombach2021HighResolutionIS} allow for manipulation of the learned latent space. We utilize such models for our dataset-conditioned weight sampling. 

\textbf{Large Language Models: } 
Large language models (LLMs)\citep{minaee2024largelanguagemodelssurvey, zhao2023surveylargelanguagemodels} like GPT-4, LLaMA\citep{dubey2024llama3herdmodels}, and Mistral have transformed natural language processing, demonstrating strong performance across diverse tasks such as summarization, translation, and text generation. While fine-tuning these models for domain-specific tasks enhances their capabilities, their large parameter count poses challenges for efficient optimization without degrading core functionality. Although weight generation offers potential for exploring optimal LLM configurations, research in this area remains limited.

\textbf{Applications of Generative Hyper-Representation Learning: } Despite skepticism around learning from pretrained weight distributions, generating diverse weights for the same model is key to improving model flexibility, initialization, and transfer learning. Task-conditioned parameter generation builds on this by encoding weight distributions from models trained on various datasets, allowing for more robust and adaptive weight retrieval, which can significantly enhance performance across tasks. While previous works, like \cite{tang2024modelgptunleashingllmscapabilities} and \cite{zhao2024retrievalaugmentedmixtureloraexperts}, focus on task-conditioned LoRA weight retrieval, and \cite{gong2024efficienttrainingdenoisedneural} explores LoRA weights for image generation, our approach extends these efforts by generating neural network weights across a wider range of applications.

% \par Our work integrates elements from the previously discussed methods, enabling dataset-conditioned neural network weight generation for known and novel datasets.

%% file: sections/approach.tex
\section{Approach}\label{sec:approach}
\subsection{Dataset conditioned Neural Networks Generation}
In this work, we first focus on the following problem: Given a set of neural network models pretrained on a collection of \( M \) datasets \(\{\mathfrak{D}_1, \mathfrak{D}_2, \ldots, \mathfrak{D}_M\}\), we aim to learn the distribution \( p(W) \) of the pretrained weights (\( W \)) of these models. Our goal is to enable conditional sampling of a set of weights \( p(W_{new}|\mathfrak{D}_{new}) \) for a new dataset or tasks \(\mathfrak{D}_{new}(x, y)\), such that these weights can achieve good performance on the new dataset either without further training or with only a few optimization steps compared to random initialization. The intuition is that there is a direct relationship between a pretrained network weights and the dataset it was trained on (see Appendix \ref{sec:former-theorem} for a formal argument). By understanding the distribution of pretrained weights and their alignment with the source dataset or tasks, we can generate high-performing weights for a target dataset with minimal or no optimization.

\subsection{Weight Encoding}\label{sec:weightenc}
\textbf{Dataset collection: } We utilize publicly available pretrained models from model zoos and additionally train target architectures on selected datasets. For each model, we extract the pretrained weights, which are then aggregated to construct our training data. The extraction process follows one one the main preprocessing methods:

\textbf{Model-wise vectorization: }For each model $\mathcal{M}_i$, we flatten the weights of each layer into a vector denoted by $w$. These vectors are concatenated to form a single vector $W_i \in \mathcal{R}^{1 \times d_i}$, where $d_i$ represents the total number of trainable parameters in the model. To ensure uniformity, all vectors are zero-padded to match the maximum length $d_{\text{max}}$ across models.

\textbf{Layer-wise vectorization} Here, instead of concatenating the weights, each layer's flattened weight vector is kept separate. This allows for layer-wise sampling during inference, treating each vectorized layer as an independent input for later stages.  Since the network layers have varying dimensions, we first zero-pad each flattened parameter vector \( \bm{w} \in \mathbb{R}^{mn} \) to a length that is a multiple of a chosen chunk size. We then split the padded vector into smaller chunks of subvectors \( \bm{\bar{w}}_i \in \mathbb{R}^{l} \), where \( i \in \{1, \dots, k\} \) and \( l = \bm{\lceil mn/k \rceil} \), ensuring that each chunk is of equal length. $mn$ represents the length of the flattened weight vectors
\par \textbf{Parametrs Encoding:} We then train a Variational Autoencoder (VAE) to encode these vectors. while minimizing the objective function defined in \ref{eqn:vae}:
\begin{equation}\label{eqn:vae}
  \mathcal{L} = - \mathbb{E}_{q_\phi(z|w)} \left[ \log p_\theta(w|z) \right]  + \beta\text{KL}\left[ q_\phi(z|x) \,||\, p(z) \right]
\end{equation}
where $w$ is the vectorized weight, $z$ is the latent representation, $p_\theta$ and $q_\phi$ the reconstruction and approximate posterior terms respectively, and $p(z)$ the prior distribution. For the prior, we used a Gaussian. $\beta$ is a fixed hyper parameters that regulate the stochasticity of the VAE. Higher value increase the randomness while lower value increases the reconstruction precision with less randomness. Model-wise and layer-wise vectorized parameters are encoded using the same VAE structure, with the only difference being in the input dimensions. 
In chunk-wise encoding, the original flattened vector \( w \) is recovered by reassembling the decoded latent chunks through concatenation. The reconstructed chunks \( \hat{w}_i \) from each layer are concatenated to ensure \( \hat{w} = \hat{w}_1 \oplus \hat{w}_2 \oplus \dots \oplus \hat{w}_k \), where \( \oplus \) denotes concatenation. And reshaping \( \hat{w} \) back into the original form \( \hat{W} \) yields a close approximation of the original weight \( W \). The quality of reconstruction is assessed by evaluating the reconstructed weights on a designated evaluation dataset or task.

\subsection{Dataset Encoding}\label{sec:denc}
\par \textbf{Image Dataset Encoding: }To sample weights conditioned on a dataset, it is crucial to establish an efficient mapping between the latent representations of pretrained weights and the datasets used for pretraining. However, encoding an entire image dataset with multiple classes and millions of samples is challenging. To overcome this, we employ the Set-based dataset encoding with the Set Transformer \citep{lee2019set}, following the approach used in dataset-adaptive methods \citep{Jeong2021TaskAdaptiveNN, lee2021rapid}, which has proven to be effective for dataset encoding. Figure~\ref{fig:setransf} in the appendix provides an overview of the dataset encoder architecture.

\par\textbf{Set-based Dataset Encoding:} Given a dataset $\mathfrak{D} = \{(x_i, y_i)\}_{i=1}^N$, where $x_i$ and $y_i$ are input-output pairs, we form subsets $s_i = (x_p^{y_i})_{p=1}^{K_i}$ with $K_i = |s_i|$ and $s_i \in \mathbb{R}^{K_i \times c \times h \times w}$, where $c$, $h$, and $w$ represent the image channel, height, and width. The dataset is reorganized into a collection of subsets $\mathcal{S} = \{s_i\}_{i=1}^C$, where $C$ is the number of distinct classes, grouping images by class. To encode the dataset, we define a transformation $\mathcal{T}$ over the subsets, mapping each set $s_i$ to an embedding $z_{s_i} \in \mathbb{R}^{1 \times d}$. Note that we reuse the notation $z$ from Section~\ref{sec:weightenc} for weight embeddings. These embeddings are aggregated into a new set $\Tilde{s}_i \in \mathbb{R}^{C \times d}$, and another transformation $\mathcal{T}$ is applied to produce the final encoded dataset $z_{\mathfrak{D}} \in \mathbb{R}^{d}$. This process is summarized as the composition of Set Transformer blocks: $z_{\mathfrak{D}} = \mathcal{T} \circ \mathcal{T}(\mathcal{S})$. In summary, we refer to this dataset encoding operation as $\mathcal{T}$. The output of the dataset encoder is independent of the number of classes and samples per class, and it does not use the labels. Training the Set Transformer modules alongside the diffusion process can be computationally expensive. To mitigate this, we propose a CLIP-style objective that aligns the dataset embeddings with the corresponding encoded pretrained weights. This alignment can be performed once and then efficiently probed linearly during the diffusion process optimization.

\par \textbf{Contrastive Dataset Encoding:} The dataset encoder is trained independently using a CLIP-style contrastive loss to align dataset embeddings with weight embeddings. This idea is similar to HyperCLIP~\citep{nava2023metalearning} for class-conditioned VQA, but instead of a text encoder, we use a dataset encoder, and the VAE encoder from Section \ref{sec:weightenc} is used to encode the corresponding weights. Specifically, we use contrastive pairs $\text{Embed}((\mathfrak{D}_i(x_j, y_j)_{j}^{c_i}, W_i)) = (z_{\mathfrak{D}_i}, z_i)$, where $z_{\mathfrak{D}_i}$ is the dataset embedding and $z_i$ is the VAE-encoded weight embedding. The Set Transformer-based dataset encoder $\mathcal{T}$ is trained to align $z_{\mathfrak{D}_i} = \mathcal{T}(\mathfrak{D}_i(x_j, y_j)_{j}^{c_i})$, where $z_{\mathfrak{D}_i} \in \mathbb{R}^{1 \times d_z}$, with the weight embeddings $z_i$ using the training objective in Eq. \ref{eq2}.
\begin{equation}
    \mathcal{L}_{CLIP}=-log\frac{exp(z_i .z_{\mathfrak{D}_i/\tau)}}{\sum_{k=0}^{N}exp(z_i.z_{\mathfrak{D}_k}/\tau)}
    \label{eq2}
\end{equation}
In this step only the frozen VAE encoder is alongside the dataset encoder. 

\par\textbf{Language Task Encoding: } To enable task-description-based parameter generation for NLP tasks, we first encode each task description using Llama-3-8B-Instruct. The output from the last hidden layer is used as the task's dataset embedding. These embeddings are then directly incorporated into the diffusion process during both training and inference.
 
\subsection{Dataset-Conditioned Parameters Generation}\label{sec:wgen}
At this stage, we have access to a pretrained VAE for encoding neural network weights and a pretrained Set Transformer module to encode entire datasets. The next stage involves defining a model to generate latent representations of weights conditioned on the dataset embeddings. We achieve this by using diffusion probabilistic models (DDPM)~\citep{NEURIPS2020_4c5bcfec, Rombach2021HighResolutionIS} trained on the latent representation of the pretrained weights..

\par\textbf{Forward Process:} Given a weight embedding $z$, obtained from the encoder of the pretrained VAE, the forward diffusion process involves successive Gaussian noise perturbations of $z$ over $T$ time steps. At time step $t$,
\begin{equation}\label{eqn:fdiff}
    p(z_t|z_{t-1})=\mathcal{N}(z_t; \mu_t=\sqrt{1-\beta_t }z_{t-1}, \beta_t I)
\end{equation}
where $\beta_t \in (0, 1)$ is the noise variance and $p(z_{1:T}|z_0)=\prod_{i=1}^{T}p(z_t|z_{t-1})$.

\par\textbf{Reverse Process:}  As in most DDPM approaches the reverse process is approximated by a neural network such that:
\begin{equation}\label{eqn:rdiff}
    p_\theta(z_{t-1} | z_t) = \mathcal{N}(z_{t-1}; \mu_\theta(z_t, t), \Sigma_\theta(z_t, t)),
\end{equation}
where $\mu_\theta$ and $\Sigma_\theta$ are neural networks.

\par \textbf{Dataset-Conditioned Training:} The diffusion model is trained on the VAE embeddings $z$, conditioned on the dataset embeddings concatenated with the latent representations of the weights. To leverage existing architectures, we designed the VAE to generate latent representations that are compatible with standard latent diffusion models with minimal adjustments, optimizing the latent diffusion objective defined in Eq. \ref{eq4}.
\begin{equation}
    \mathcal{L}_{LDM}=\mathbb{E}_{z,\varepsilon \sim \mathcal{N}(0, 1),Z_{\mathfrak{D}},t}\left[||\varepsilon-\varepsilon_{\psi}(z_t,z_{\mathfrak{D}}, t)||^2_2\right]
    ,\label{eq4}
\end{equation}
where $\varepsilon_{\psi}(z_t,z_{\mathfrak{D}},t)$ is implemented as a UNet.

\par\textbf{Sampling:} New weights are sampled conditionally through the reverse diffusion process as follows:
\begin{equation}
    z_t= \frac{1}{\sqrt{a_t}}(z_t-\frac{\beta_t}{\sqrt{1-\Tilde{a}_t}}\varepsilon_{\psi}(z_t, z_{\mathfrak{D}}, t, ))+\sigma\xi
    , \label{eqr}
\end{equation}
where $\xi \sim \mathcal{N}(0, I)$ and, $\sigma_t $ a chosen value. After sampling a latent representation $\bar(z$ for a given dataset $\mathfrak{D}_i)$. The pretrained VAE decoder is used to transform these latents into a weight vector $\bar{w}= \mathcal{D}(\bar{z}) $, which is then used to initialize the target network as shown in Figure~\ref{fig:overview}.

\subsection{Exploring the Optimal Parameters Space of LLMs}
Our goal is to enhance the performance of pre-trained LLMs on a target dataset \textit{without direct fine-tuning}. To accomplish this, we introduce layer-conditioned \ourmethod, which explores optimal model parameters in latent space. The idea is to sample multiple sets of weights and identify those that improve performance on the target dataset, starting from a single pre-trained model, with no additional training. One of the key challenges in generating LLM parameters is learning the distribution of their vast number of parameters, which is computationally expensive and impractical. Therefore, it is crucial to identify a subset of parameters that, when optimized, leads to significant task performance improvements. To tackle this, we adopt the spectrum technique~\citep{hartford2024spectrumtargetedtrainingsignal}, which ranks LLM layers based on their importance, offering an efficient approach to achieving substantial performance gains without updating all parameters. We provide more detailed in the appendix in \ref{app:layer_selection} and \ref{app:approach}

% \textbf{Layer Selection}: In our evaluation, we prioritize layers with higher signal-to-noise ratios (SNR) by focusing our analysis on the top 25\% of layers

%% file: sections/experiments.tex
\input{tables/meta_results}

\input{tables/zeroshot_swint}
\section{Experiments}\label{exps}
We present two sets of results: weight generation with and without finetuning. For weight generation without finetuning, we present results on Few-Shot Learning, Zero-Shot Learning and Model Retrieval in Section~\ref{exp:nofinetuning}. In Section~\ref{exp:finetune}, we present a set of results where the generated weights are further finetuned. For all experiments, we utilize a single Titan RTX GPU with 24GB of memory. An extensive set of ablation studies on the proposed method is provided in Appendix~\ref{app:ablation} and \ref{recap}

\subsection{Weight Generation without Finetuning}\label{exp:nofinetuning}
We present a set of results where the generated weights are evaluated directly without finetuning for few-shot learning, zero-shot learning and model retrieval.
\subsubsection{Weights Generation for Few-Shot Learning}
\par\textbf{Task:}  We aim to show that learning the distributions of model pretrained independently on a large set of dataset can enable sampling weights that compete with meta-learning techniques in multi-task few-shot learning, without requiring fine-tuning. 

\par\textbf{Dataset:} We utilize the \textit{mini}-ImageNet and \textit{tiered}-ImageNet datasets for this task. For the architectures, we use a four-layer ConvNet and a ResNet12 backbone provided by \cite{chen2021meta}. We generate the pretrained weights by linear probing a classifier head on each of the 50,000 subsets for 10 epochs and evaluate the performance on 600 subsets from the unseen test split for 1-shot and 5-shot. Analogously to few shot learning, we choose the number of images per class for conditioning to be the same as the support set, while the number of images per class in the query set is fixed to 15 for all methods and 600 tasks are used for testing.

\par\textbf{Baselines:} We benchmark against iMAML~\citep{NEURIPS2019_072b030b}, ALFA~\citep{NEURIPS2020_ee89223a}, COMNL~\citep{deleu2022comln}, MetaQDA~\citep{9710819}, MetaDiff~\citep{Zhang_2024}, MetaOptNet~\citep{8954109} and a classifier baseline introduced in \citet{chen2021meta}.

\par\textbf{Results:} Table~\ref{table_meta} shows that our approach consistently improves performance on all tasks while utilizing the same backbone as other methods. With the Conv4 backbone, we achieve approximately $6\%$ performance improvement in 1-shot learning and 3 to 4\% on 5-shot learning on mini-ImageNet. On Tiered-ImageNet, we achieve more than 8\% performance improvement on 1-shot and 5 to 6\% average improvement on 5-shots.  For the ResNet12 backbone we achieve 4 to 9\% performance improvement. These results demonstrate the effectiveness of our method against the existing meta-learning methods.

\par We follow the standard image classification setup with and sample 50 weights for each subset, averaging the top 3 accuracies of all subsets as outlined in Table \ref{table_meta}. 
In 1-shot learning, conditioning is based on a single image per class per dataset, whereas in 5-shot learning, conditioning relies on five images per class per dataset. Dataset-conditioned weight generation aligns with meta-learning by enabling rapid model adaptation to new tasks. The generated weights are tailored to specific dataset features, allowing the model to leverage prior knowledge and achieve improved generalization surpassing meta-learning approaches.

\subsubsection{Zero-Shot Classifier Head Adaptation}\label{zerodapt}
\par\textbf{Task:} We evaluate the performance of the proposed method in adapting the classifier head to unseen datasets. 
%Large models pretrained on datasets like ImageNet are frequently used as feature extractors, with only the classifier head fine-tuned for new tasks. 
In this experiment, we assess whether our method can conditionally generate the classifier weights, potentially eliminating or significantly speeding up the finetuning process.

\par\textbf{Dataset:} We partitioned ImageNet-1k into 20k subsets 
of 50 classes each with 50 images per class per subset and linear probe a classifier head for 10 epochs using Tiny Swin Transformer (denoted Swin in Table~\ref{table_meta}), and 
ResNet18 all pretrained on ImageNet-1k. For dataset conditioning, we use 5 images per class per subset. The unseen target datasets are CIFAR-10, STL-10, Aircraft, Pets, and CIFAR-100 . The baseline methods in these experiments are ResNet18 and Tiny Swin Transformer pretrained on ImageNet-1k.

\par\textbf{Baselines:} We benchmark against the pretrained backbones, and two GHN models~\citep{knyazev2021parameter,knyazev2023canwescale}. Additonally, we provide a powerful variant of our model D2NWG\_CLIP where the dataset encoder encodes the CLIP embedding for each sample in the datasets.

\par\textbf{Results:} Table~\ref{tab:swin} presents the performance of the sampled weights where it can be seen that the proposed method achieves better performance compared to the ImageNet pretrained weights and the GHN family of models. Additionally, the variant of our model that utilizes the CLIP embedding for dataset encoding significantly improves the performance suggesting that better dataset representation learning can boost the performance of the generated weights.

\input{tables/augmeted_retriever}

\subsubsection{Model Retrieval}\label{aug}
\par\textbf{Task:}  We assess the Generative Augmented Retrieval capability of \ourmethod, aiming to show that it can learn the distribution of models pretrained on diverse real-world datasets. This task requires generation of dataset-conditioned weights that achieve performance comparable to the original pretrained models and hence provide access to a wide range of pretrained models through efficient sampling.

\par\textbf{Dataset:} We collected 30 real-world datasets\citep{meta-album-2022}, spanning 19 to 706 classes and organised into 10 domains with 3 datasets per domain, and fine-tuned a MobileNetV3 subnet\footnote{\url{https://pytorch.org/hub/pytorch_vision_once_for_all/}}
sampled from OFA~\citep{cai2020once} for 100 epochs on each dataset. We then learned the distribution of the combined pretrained models from the last 20 epochs across all datasets.

\par\textbf{Baselines:} For this task, we compare with the original pretrained weights which are finetuned on each individual dataset. For each dataset, we sample and report the average accuracy of 5 set of weights sampled with \ourmethod.

\textbf{Results: } From Table~\ref{tab:augmet} we see that \ourmethod conditionally generates high-performing parameters while enhancing the pretrained model, achieving the best average results across all datasets. This demonstrates the strong retrieval capability of our method, suggesting it can be used as a neural network weight retriever in approaches like~\citep{zhao2024retrievalaugmentedmixtureloraexperts}, eliminating the need for pretrained database. Detailed dataset information is provided in Table \ref{tab:dataset_table} and more experiments in the Appendix \ref{app:full}. Additionally, it is much more efficient to generate weights with our model compared to pretraining as shown by the runtime in Table~\ref{tab:augmet}.

% \input{figures/vit_bar}

%\par \textbf{Evaluation with Full Vision transformer}: Our method shows the ability to learn the distribution of all parameters within a vision transformer, including convolutional and linear layers. We present in-distribution evaluation results, highlighting the learning of combined weight distributions conditioned on individual datasets. Additional findings on generating full model weights are detailed in the appendix.

\input{tables/hyperzooTable}

\subsection{Weight Generation with Fine-tuning}\label{exp:finetune}
In this section, we evaluate the quality of the generated weights in fine-tuning scenarios to assess their suitability for transfer learning.

\textbf{Task: } The goal is to assess the behavior of the sampled weights when finetuned on the same dataset and compare convergence speed. This experiment focuses on evaluating whether the sampled weights can be effectively fine-tuned to achieved superior final performance, rather than simply aiming for weights producing high initial accuracy and may not lead to superior performance while fine-tuning.

\textbf{Datasets}: We used the modelzoo of \citet{schurholtModelZoosDataset2022} consisting of a ConvNet trained on MNIST, SVHN, CIFAR-10 and STL-10. Our model was trained using the combined pretrained weights from epochs 21 to 25 of all models, consistent with the baseline settings.

\textbf{Baselines:} We compare against the kernel density estimator approaches from \citet{schuerholt2024sane, Schrholt2022HyperRepresentationsAG}, evaluated on the same datasets. Unlike these unconditional methods, we build a model specifically for MNIST and SVHN, and another for CIFAR-10 and STL-10. For each dataset, five sets of weights were sampled to initialize the models, which were fine-tuned for a number of epochs from 0 to 25. We also add RandomInit model trained for 50 epochs and show that our sampled weight finetuned for 25 epochs outperforms this model.

\textbf{Results: } As shown in Table~\ref{tab:hyperzoo}, \ourmethod consistently accelerates convergence across related tasks, surpassing the pretrained model and outperforming both baselines ~\citet{schurholtHyperRepresentationsGenerativeModels2022, schuerholt2024sane}. The finding implies that \ourmethod accelerates convergence and improves performance compared to existing methods. This highlights its potential for faster and more efficient model initialization, making it valuable for transfer learning and real-world applications. Interestingly, on MNIST and SVHN, weights with higher initial performance tend to degrade during fine-tuning.

\subsection{Performance Evaluation on Unseen Datasets}
\textbf{Task:} The objective remains the same as in Section~\ref{exp:finetune}, but here we evaluate the proposed method solely on unseen datasets.

\textbf{Datasets:} We assess \ourmethod on a real-world dataset of 140 subsets with class counts ranging from 2 to 20, and 10 test sets with up to 1,566 classes. We use a two-layer MLP on top of a CLIP image encoder and fine-tune it on training datasets  to collect the pretrained zoo.(see appendix \ref{zoos}). The image datasets contains food, dataset, drawing, x-ray and others.

\textbf{Baselines:} The baseline methods are random initialization and a pretrained MLP previously trained on ImageNet.

\textbf{Results:} Figure~\ref{fig:kaggle} shows performance on four unseen datasets, where \ourmethod achieves 99.04\% initial accuracy on the dessert dataset, outperforming the randomly initialized model even after 50 epochs. \ourmethod consistently accelerates convergence across all tasks, surpassing both random and pretrained initialization. Despite no class overlap between training and test datasets, it demonstrates strong transferability. Detailed results are available in Table~\ref{tab22} of the Appendix.

\input{figures/best_kaggle}

\subsection{Transfer Learning: MobileNet Full Weight Generation}

\textbf{Task:} We evaluate each method's generalization on CIFAR-10, STL-10, Pets and Aircrafts, focusing on performance gains in domain-specific tasks. The goal is to identify the best initialization strategy for improving model adaptability across diverse data distributions.
\input{figures/transfer}

\textbf{Baseline:} The baseline in this experiment are the Pretrained model, which uses weights from a model pretrained on ImageNet and RandomInit, a randomly initialized model. %Each model is fine-tuned with Adam optimizer with lr=1e-3 and weight gecay set to 1e-3.

\textbf{Datasets:} In this experiment we evaluate the transferability to unseen dataset of \ourmethod trained in Section~\ref{aug} on unseen datasets CIFAR-10, STL-10, Aircraft100, Aircraft30, and Pets.

\textbf{Results:} We initialized the target model with 5 sampled weights and fine-tuned it for 1 epoch on each dataset, along with 5 pretrained and randomly initialized models. Results are provided in Figure~\ref{bartransf} where we transfer across CIFAR-10, STL-10, Aircraft100, Aircraft30, and Pets and show that D2NWG outperforming the baselines. On the AIRCRAFT-100 dataset, our method achieved an accuracy of $1.43\%$, outperforming both the randomly initialized model, which reached 1.0\%, and the ImageNet-pretrained model, which achieved $1.24\%$. D2NWG demonstrates superior generalization and adaptability, making it a more effective initialization strategy even on specialized datasets such as Aircrafts and Pets.

\input{tables/loraTable}

\vspace{-0.15in}
\subsection{Task Conditioned LoRA weights Generation}
\textbf{Task: } In this section, we demonstrate that our method can be applied to LLMs by learning the distribution of LoRA matrices conditioned on task-specific textual descriptions. %, as the pretrained dataset is also textual.

\textbf{Datasets: } We use six tasks from the GLUE benchmark and generate task descriptions using GPT-4, as shown in Table \ref{tab:glues_descriptor}. LoRA weights were generated following the fine-tuning process of \citet{gao2024parameterefficientfinetuningdiscretefourier}. We collected LoRA and classifier head checkpoints from the last 5 epochs, combined the pretrained vectors, and conditionally learned their distribution.

\textbf{Baselines: } We compare  with base Roberta-base, LoRA~\citep{lora}, AdaLoRA~\citep{adalora}, DyLoRA~\citep{dylora} and FourierFT~\citep{gao2024parameterefficientfinetuningdiscretefourier} which are all LoRA-based RoBERTa-base models. We sampled and compared the average accuracy of the top 5 performing sets of weights per dataset.

\textbf{Results: }As shown in Table \ref{lora}, \ourmethod effectively generates weights that match or surpass the performance of pretrained models. These results align with our findings from the augmented weight retrieval experiments. Additional details regarding the task descriptor are provided in Table \ref{tab:glues_descriptor}.
% tab:glues_descriptor
\input{tables/llm_task_specific}
\subsection{Enhancing LLM Performance with Weight Sampling}
\textbf{Task: } We aim to demonstrate that \ourmethod can enhance existing LLMs by learning the distribution of their pretrained weights, enabling the generation of parameters that improve performance on specific tasks while generalizing to unseen tasks.

\textbf{Datasets: } We evaluate on several benchmarks\citep{open-llm-leaderboard}: AI2 Reasoning Challenge (25-shot) for grade-school science questions, HellaSwag (25-shot) for commonsense inference, Winogrande (5-shot) for commonsense reasoning.

\textbf{Baseline: } We evaluate our method against various version of LLAMA3 and Mistral-7B.

For each model, We extract the weights of the top 25\% of layer excluding embedding and output layer, learn their distribution using chunk based encoding, We then steer through the optimal space to generate task-specific parameters as shown in Table \ref{tab:task-spec}.

\textbf{Results: } The results in Table \ref{tab:task-spec} demonstrates that our approach consistently improve the performance of each models demonstrating new application avenues of our proposed method. 

\subsection{Evaluation on Open LM Benchmark}
We  merge these models following \citet{pmlr-v162-wortsman22a} and evaluate them on the OpenLM leaderboard \citep{open-llm-leaderboard-v2} as shown in Table \ref{tabgenerator}

\textbf{Task: }We evaluate the robustnets of ours best models on the open-lm leaderboard.

\textbf{Datasets: } We evaluate models on 6 key benchmarks using the Eleuther AI Language Model Evaluation Harness: IFEval (0-shot) for instruction adherence, BBH (Big Bench Hard, 0-shot) with 23 challenging tasks (arithmetic, reasoning, language understanding), MATH (0-shot) focusing on Level 5 high-school math problems, GPQA (0-shot) with graduate-level Q\&A across various fields, MuSR (0-shot) testing complex reasoning with long-range context, and MMLU-Pro (0-shot) for advanced multitask knowledge assessment. These benchmarks assess diverse reasoning and knowledge capabilities in 0-shot and few-shot settings.

\textbf{Baselines:} We compare our method against LLMA3.1-8B-Instruct and its fine-tuned variant, with evaluations conducted on the leaderboard server.

\textbf{Results:} As shown in Table\ref{leader_board} , our method outperforms the base models on the leaderboard and is on par with models pretrained on task-specific datasets. Although the weights were not directly calibrated using the tasks from the leaderboard, \ourmethod achieves up to a 3\% improvement on some tasks. This confirms that it is possible to guide a model towards specialization in certain tasks by effectively exploring the optimal parameter space through sampling. The consistent performance gains across diverse benchmarks underscore D2NWG’s effectiveness in improving model robustness and transferability. Ours llama-3.2-1B based model\footnote{\url{https://huggingface.co/DeepAutoAI/Explore_Llama-3.2-1B-Inst_v1.1}} ranked among the top llama-3.2-1B models on the public leaderboard.

\textbf{Quality Check:} Additionally, our method improves text generation quality, as demonstrated by the experimental results in Table \ref{tabgenerator}. We demonstrate that the generated weights are capable of generating text on par with the base pretrained model. The model used for quality check ranked 4\textsuperscript{th} on open lm-leaderboard\footnote{ \url{https://huggingface.co/DeepAutoAI/Explore_Llama-3.1-8B-Inst}}. Further results on LLMs are provided in Appendix \ref{llm2}, where we also demonstrate that the proposed method successfully learns the full parameters of GPT-2 small while maintaining performance comparable to the pretrained model.

 \input{tables/llm_leaderboard}

\par \textbf{Limitations}: Our method relies on large collections of pretrained weight tensors and datasets, which require substantial storage and computational resources. However, such pretrained models are becoming more readily available due to the efforts made by open-source communities.

%% file: tables/meta_results.tex
\begin{table*}[t]
\caption{Few-Shot Learning. ALL implies generation of the entire parameters and CH denotes generation of classification head only. }
\vspace{-0.2in}
	\begin{center}
		\resizebox{1.0\columnwidth}{!}{
			\begin{tabular}{l|c|c|c|c|c|c}
				\toprule
				%\hline
				\multicolumn{1}{l|}{\multirow{2}{*}{Method}}&\multicolumn{1}{c|}{\multirow{2}{*}{Adaptation}}&\multicolumn{1}{c|}{\multirow{2}{*}{Backbone}}& \multicolumn{2}{c|}{\textit{mini}-ImageNet} & \multicolumn{2}{c}{\textit{tiered}-ImageNet} \\ 
				\cline{4-7}
				& & & 5-way 1-shot & 5-way 5-shot & 5-way 1-shot & 5-way 5-shot \\
				%\hline
				\hline
			
				iMAML~\citep{NEURIPS2019_072b030b} & ALL & Conv4 & $49.30 \pm 1.88\%$  & $59.77 \pm 0.73\%$ & $38.54 \pm 1.37\%$  & $60.24 \pm 0.76\%$ \\
				
				ALFA ~\citep{NEURIPS2020_ee89223a} & ALL & Conv4 & $50.58 \pm 0.51\%$  & $69.12 \pm 0.47\%$ & $53.16 \pm 0.49\%$  & $70.54 \pm 0.46\%$ \\
		
				COMLN ~\citep{deleu2022comln} & CH & Conv4 & $53.01 \pm 0.62\%$  & $70.54 \pm 0.54\%$ & $54.30 \pm 0.69\%$  & $71.35 \pm 0.57\%$ \\ 
				MetaQDA ~\citep{9710819} & CH & Conv4 & $56.41 \pm 0.80\%$  & $72.64 \pm 0.62\%$ & $58.11 \pm 0.48\%$  & $74.28 \pm 0.73\%$ \\ 
				%\midrule
				% \hline
				MetaDiff ~\citep{Zhang_2024} & CH & Conv4 & 55.06 $\pm$ 0.81$\%$ & 73.18 $\pm$ 0.64$\%$ & 57.77 $\pm$ 0.90$\%$ & 75.46 $\pm$ 0.69$\%$  \\	
                %\hline	
                D2NWG(Ours) & CH & Conv4 & \textbf{61.13} $\pm$ \textbf{8.50}$\%$ & \textbf{76.94} $\pm$ \textbf{6.04}$\%$ & \textbf{65.33} $\pm$ \textbf{6.50}$\%$ & \textbf{80.05} $\pm$ \textbf{8.25}$\%$ \\	
				
				\midrule 
    % \hline

				ALFA ~\citep{NEURIPS2020_ee89223a} & ALL & ResNet12 & $59.74 \pm 0.49\%$  & $77.96 \pm 0.41\%$ & $64.62 \pm 0.49\%$  & $82.48 \pm 0.38\%$ \\
			
				MetaOptNet~\citep{8954109} & CH & ResNet12 &  $62.64 \pm 0.61\%$  & $78.63 \pm 0.46\%$ & $65.99 \pm 0.72\%$  & $81.56 \pm 0.53\%$ \\
				LEO ~\citep{rusu2018metalearning} & CH & WRN-28-10 & $61.76 \pm 0.08\%$  & $77.59 \pm 0.12\%$ & $66.33 \pm 0.05\%$  & $81.44 \pm 0.09\%$ \\
		
				Classifier ~\citep{chen2021meta} & CH & ResNet12 & 61.22 $\pm$ 0.84$\%$ & 78.72 $\pm$ 0.60$\%$ & 69.71 $\pm$ 0.88$\%$ & 83.87 $\pm$ 0.64$\%$ \\
				MetaQDA ~\citep{9710819} & CH & ResNet18 & 65.12 $\pm$ 0.66$\%$ & 80.98 $\pm$ 0.75$\%$ & 69.97 $\pm$ 0.52$\%$ & 85.51 $\pm$ 0.58$\%$ \\
				%\midrule
				% \hline
				MetaDiff ~\citep{Zhang_2024} & CH & ResNet12 & 64.99 $\pm$ 0.77$\%$ & 81.21 $\pm$ 0.56$\%$ & 72.33 $\pm$ 0.92$\%$ & 86.31 $\pm$ 0.62$\%$ \\	
				%\hline		

                    D2NWG(Ours) & CH & ResNet12 & \textbf{69.55}$\pm$ \textbf{3.77}$\%$ & \textbf{83.51} $\pm$ \textbf{6.21}$\%$ & \textbf{81.15} $\pm$ \textbf{9.70}$\%$ & \textbf{90.04} $\pm$ \textbf{6.10}$\%$ \\	
				%\hline	
                    \bottomrule
		\end{tabular}}
	\end{center}
	\label{table_meta}
 \vspace{-0.1in}
\end{table*}

%% file: tables/zeroshot_swint.tex
\begin{table*}[t]
\caption{\small Zero-Shot Transfer Learning. We evalutate on two backbones: Tiny Swin Transformer and ResNet18.}% This Table represents the results of classifier head adaptation using swin transformer, clip, and resnet18 classifier head. The results on Aircraft, Pets, and CIFAR-100 represent the average accuracy of each dataset divided into disjoint subsets containing a maximum of 10 classes.}
\vspace{-0.2in}
\begin{center}
\resizebox{0.8\columnwidth}{!}{
\begin{tabular}{lccccc}
\toprule
Model & CIFAR-10& STL-10 & Aircraft & Pets& CIFAR-100\\
\midrule
Swin & 7.38 & 8.43 & 5.01  & 2.63 &1.35\\
%GHN2~\citep{knyazev2021parameter} &  48.20& -- & --  & --  &12.7\\
%GHN3~\citep{knyazev2023canwescale} & 51.8 & -- & --  & --  &11.9\\
GHN2~\citep{knyazev2021parameter} &  48.20& -- & --  & --  &12.7\\
GHN3~\citep{knyazev2023canwescale} & 51.8 & -- & --  & --  &11.9\\
D2NWG(Ours) & \textbf{53.12 $\pm$ 0.25} & \textbf{60.42 $\pm$ 0.14} & \textbf{24.57 $\pm$ 3.16}  & \textbf{26.47 $\pm$ 1.90}& \textbf{30.44 $\pm$ 0.15}\\
\midrule
ResNet18 &  10.88& 6.78& 3.75 & 2.39 &1.38\\
GHN2~\citep{knyazev2021parameter} &  19.52& 13.04 & --  & --  &--\\
D2NWG & 33.03 $\pm$ 0.04 & 50.42 $\pm$ 0.13 & 17.60 $\pm$ 2.13  &17.29$\pm$ 0.13& 13.71 $\pm$ 0.63\\
D2NWG\_CLIP(Ours) & \textbf{60.42$\pm$ 0.75} & \textbf{82.42 $\pm$ 0.04} & \textbf{27.70 $\pm$3.24}  & \textbf{32.17 $\pm$6.30}  & \textbf{51.50 $\pm$0.25}\\

\bottomrule
\end{tabular}			
   }
\end{center}
\label{tab:swin}
\vspace{-0.2in}
\end{table*}

%% file: tables/augmeted_retriever.tex
\begin{wrapfigure}{t}{0.4\textwidth}
\small
\centering
\vspace{-0.3in}
\captionof{table}{\small Model Retrieval via Generative Augmented Weight Sampling}%on 30 Real-World Datasets. %\textbf{Runtime} Time required to generate the optimal parameters}
\vspace{-0.1in}
\resizebox{0.4\textwidth}{!}{
\begin{tabular}{lcc}
\toprule
Domain & Pretrained & \ourmethod(Ours) \\
\midrule
Large Animals    & 71.11 $\pm$ 11.45  & 70.33 $\pm$ 12.42 \\
Small Animals    & 54.04 $\pm$ 13.56  & 54.70 $\pm$ 13.83 \\
Plants           & 63.69 $\pm$ 9.05   & 71.37 $\pm$ 17.15 \\
Plant Diseases   & 81.69 $\pm$ 19.14  & 81.98 $\pm$ 19.53 \\
Microscopy       & 55.56 $\pm$ 26.14  & 55.49 $\pm$ 26.17 \\
Remote Sensing   & 82.20 $\pm$ 7.49   & 82.68 $\pm$ 8.05 \\
Vehicles         & 57.07 $\pm$ 19.57  & 58.09 $\pm$ 18.30 \\
Manufacturing    & 84.34 $\pm$ 21.00  & 84.32 $\pm$ 20.96 \\
Human Actions    & 68.63 $\pm$ 12.45  & 69.09 $\pm$ 12.73 \\
OCR              & 63.18 $\pm$ 1.75   & 65.60 $\pm$ 2.00  \\ 
\midrule
Average & 68.32 $\pm$ 13.84 & \textbf{69.47 $\pm$ 14.79} \\
Runtime&6 hours& 40 seconds\\

\bottomrule
\end{tabular}
}
\label{tab:augmet}
\vspace{-0.1in}
\end{wrapfigure}

%% file: tables/hyperzooTable.tex
\begin{wrapfigure}{t}{0.5\textwidth}
\small
\centering
\vspace{-0.15in}
\captionof{table}{\small Finetuning of Generated Weights using the Modelzoo of \citet{schurholtModelZoosDataset2022}.}
%Models weights generated fine-tuned on the same pretrained tasks. We compare training from scratch with $S_{KDE30}$ from \citep{Schrholt2022HyperRepresentationsAG}, and $SANE$ from \citep{schuerholt2024sane} combined with the $KDE30$ sampling method, and their $SANE_{SUB}$  subsampled to \ourmethod. }
\vspace{-0.1in}
\resizebox{0.5\textwidth}{!}{
\begin{tabular}{clcccc}
\toprule
Epoch                & Method & MNIST               & SVHN               & CIFAR-10           & STL                \\
\hline

0 & RandomInit                  & $\sim$10 /\%        & $\sim$10 /\%       & $\sim$10 /\%       & $\sim$10 /\%       \\
0 & $S_{KDE30}$                & 68.6$\pm$6.7           & 54.5$\pm$5.9          & \textit{n/a}                & \textit{n/a}                \\
0 & $SANE_{KDE30}$                    & 84.8$\pm$0.8           & 70.7$\pm$1.4          & 56.3$\pm$0.5          & 39.2$\pm$0.8          \\
0 & $SANE_{SUB}$                   & \textbf{86.7$\pm$0.8}  & \textbf{72.3$\pm$1.6} & 57.9$\pm$0.2 & 43.5$\pm$1.0 \\
0 & \ourmethod                 & 80.52$\pm$0.82                & 66.6$\pm$0.7    & \textbf{58.80$\pm$0.1}  & \textbf{44.50$\pm$0.1}              \\

\midrule

1 & RandomInit                  & 20.6$\pm$1.6           & 19.4$\pm$0.6          & 37.2$\pm$1.4          & 21.3$\pm$1.6          \\
1 & $S_{KDE30}$                & 83.7$\pm$1.3           & 69.9$\pm$1.6          & \textit{n/a}                & \textit{n/a}                \\
1 & $SANE_{KDE30}$                    & 85.5$\pm$0.8           & 71.3$\pm$1.4          & 58.2$\pm$0.2          & 43.5$\pm$0.7          \\
1 & $SANE_{SUB}$                   & 87.5$\pm$0.6  & 73.3$\pm$1.4 & 59.1$\pm$0.3 & 44.3$\pm$1.0 \\
1 & \ourmethod                 & \textbf{87.8$\pm$0.4} & \textbf{73.6$\pm$1.3}  & \textbf{59.2$\pm$0.3}       & \textbf{44.8$\pm$0.2}              \\

\midrule

5 & RandomInit                  & 36.7$\pm$5.2           & 23.5$\pm$4.7          & 48.5$\pm$1.0          & 31.6$\pm$4.2          \\
5 & $S_{KDE30}$                & 92.4$\pm$0.7 & 57.3$\pm$12.4         & \textit{n/a}                & \textit{n/a}                \\
5 & $SANE_{KDE30}$                    & 87.5$\pm$0.7           & 72.2$\pm$1.2          & 58.8$\pm$0.4          & 45.2$\pm$0.6          \\
5 & $SANE_{SUB}$                   & 89.0$\pm$0.4           & 73.6$\pm$1.5 & 59.6$\pm$0.3 & 45.3$\pm$0.9 \\
5 & \ourmethod                 & \textbf{92.5$\pm$0.9}  & \textbf{74.0$\pm$0.1}  & \textbf{60.3$\pm$0.1}      & \textbf{45.4$\pm$0.1}               \\

\midrule

25 & RandomInit                  & 83.3$\pm$2.6           & 66.7$\pm$8.5          & 57.2$\pm$0.8          & 44.0$\pm$1.0          \\
25 & $S_{KDE30}$                & 93.0$\pm$0.7           & 74.2$\pm$1.4          &       \textit{n/a}             &    \textit{n/a}       \\
25 & $SANE_{KDE30}$                    & 92.0$\pm$0.3           & 74.7$\pm$0.8          & 60.2$\pm$0.6          & 48.4$\pm$0.5 \\
25 & $SANE_{SUB}$                   & 92.3$\pm$0.4           & 75.1$\pm$1.0 & 61.2$\pm$0.1 & 48.0$\pm$0.4          \\
25 & \ourmethod                 & \textbf{96.2$\pm$0.3}       & \textbf{75.7$\pm$0.5 } & \textbf{64.1$\pm$1.0}    & \textbf{48.7$\pm$0.5}        \\

\midrule
50 & RandomInit                  & 91.1$\pm$2.6           & 70.7$\pm$8.8          & 61.5$\pm$0.7          & 47.4$\pm$0.9      \\

\bottomrule
\end{tabular}
}
\label{tab:hyperzoo}
\vspace{-0.1in}
\end{wrapfigure}

%% file: figures/best_kaggle.tex
\begin{figure*}[t!]
 \vspace{-0.15in}
\centering
    \begin{subfigure}{0.24\textwidth}
        \centering
        \includegraphics[width=\linewidth]{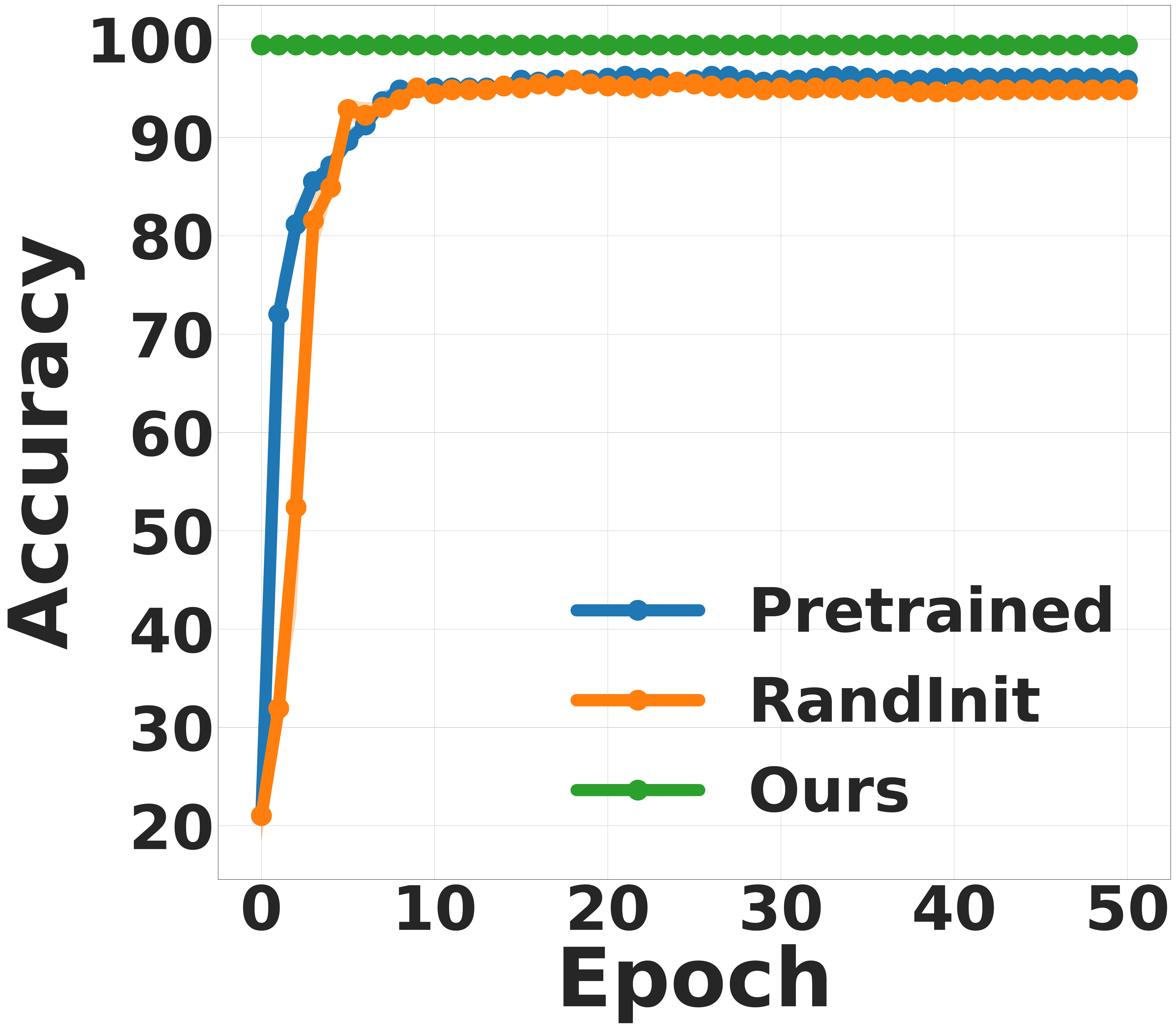}
        \captionsetup{justification=centering,margin=0.5cm}
        \caption{Dessert Food Dataset}
    \end{subfigure} 
    \begin{subfigure}{0.24\textwidth}
	\centering
	\includegraphics[width=\linewidth]{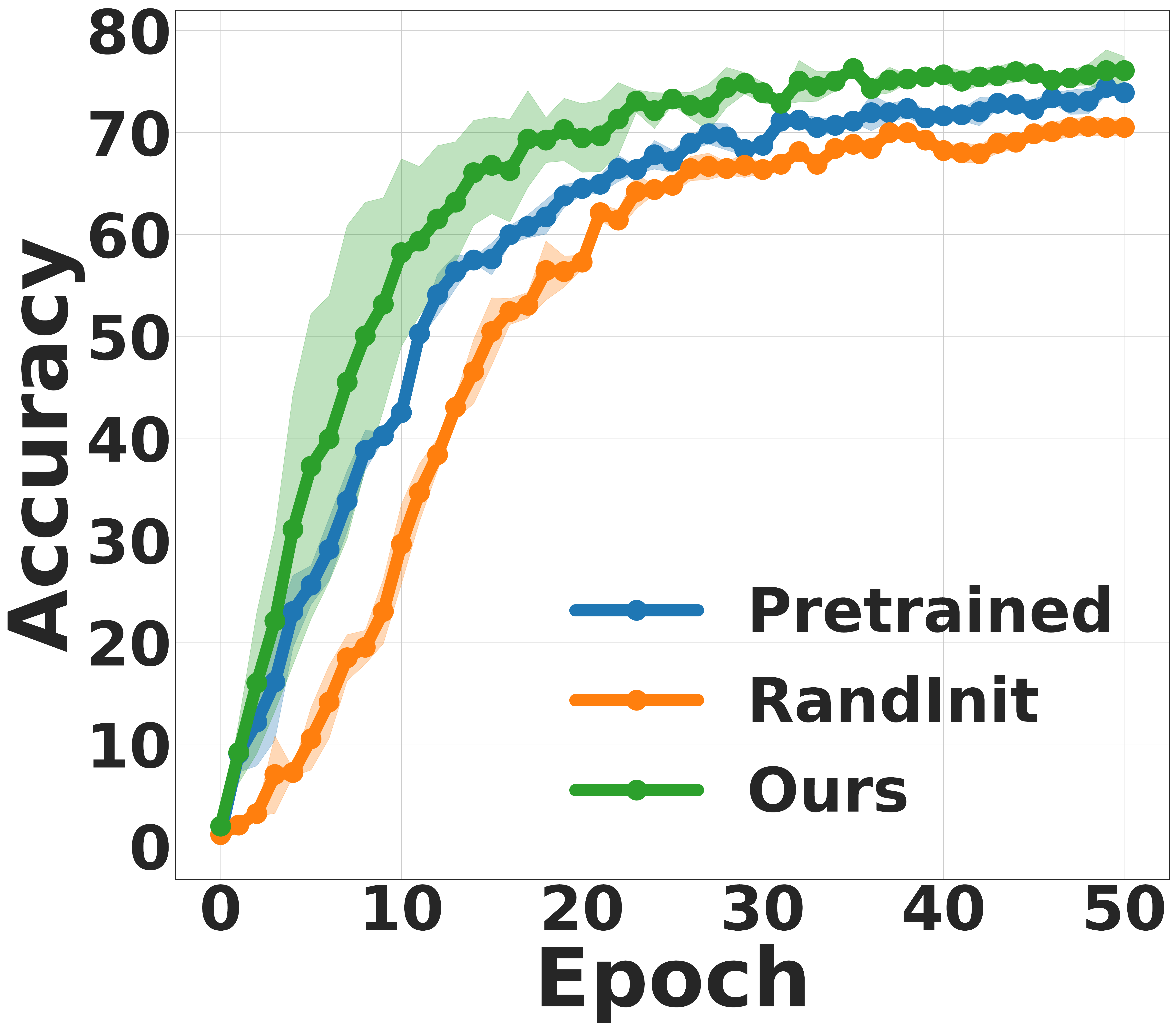}
	\captionsetup{justification=centering,margin=0.5cm}
	\caption{Gemstones Dataset}
    \end{subfigure}
    \begin{subfigure}{0.24\textwidth}
	\centering
	\includegraphics[width=\linewidth]{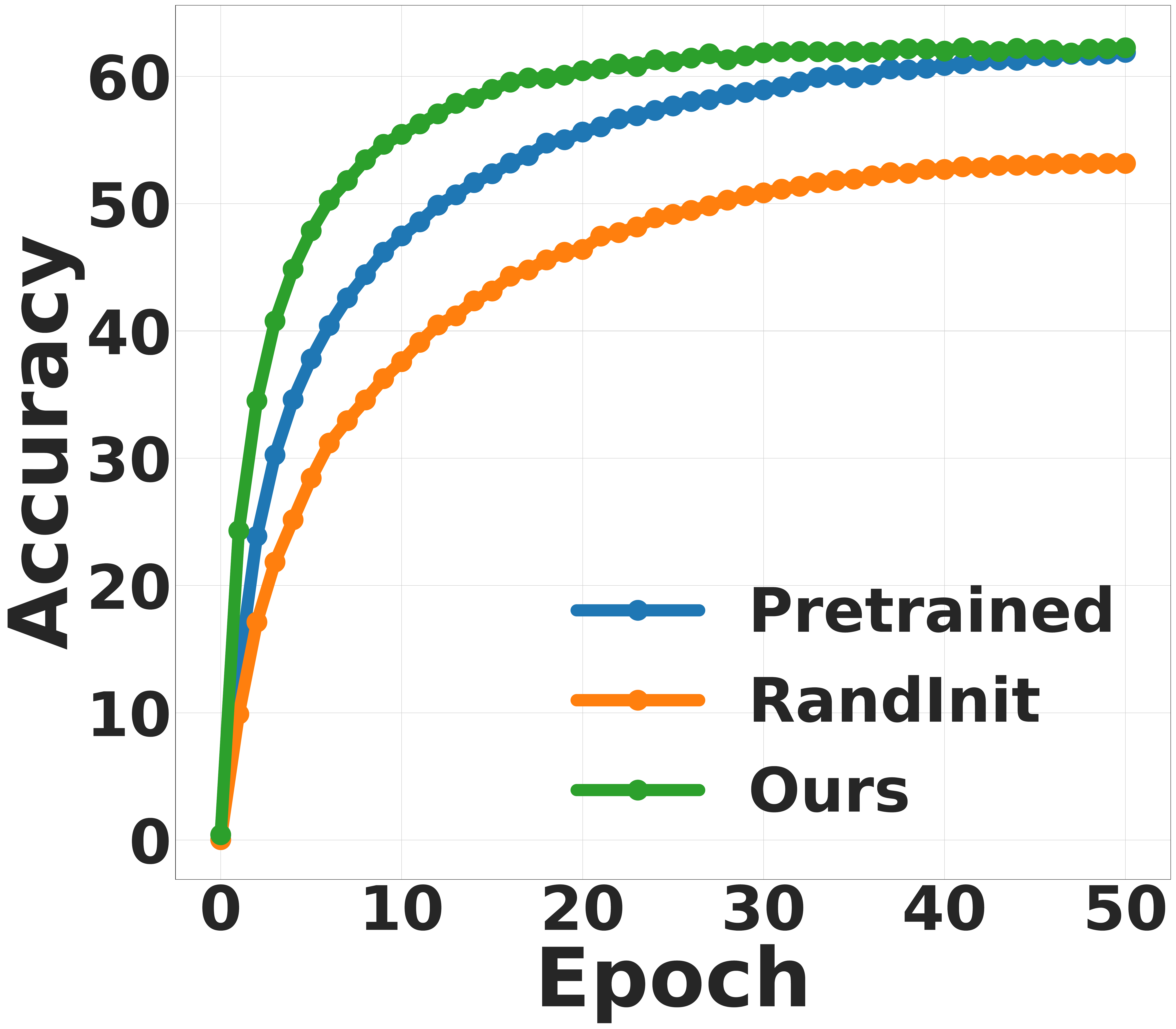}
	\captionsetup{justification=centering,margin=0.5cm}
	\caption{Japanese Characters}
    \end{subfigure}
    \begin{subfigure}{0.24\textwidth}
	\centering
	\includegraphics[width=\linewidth]{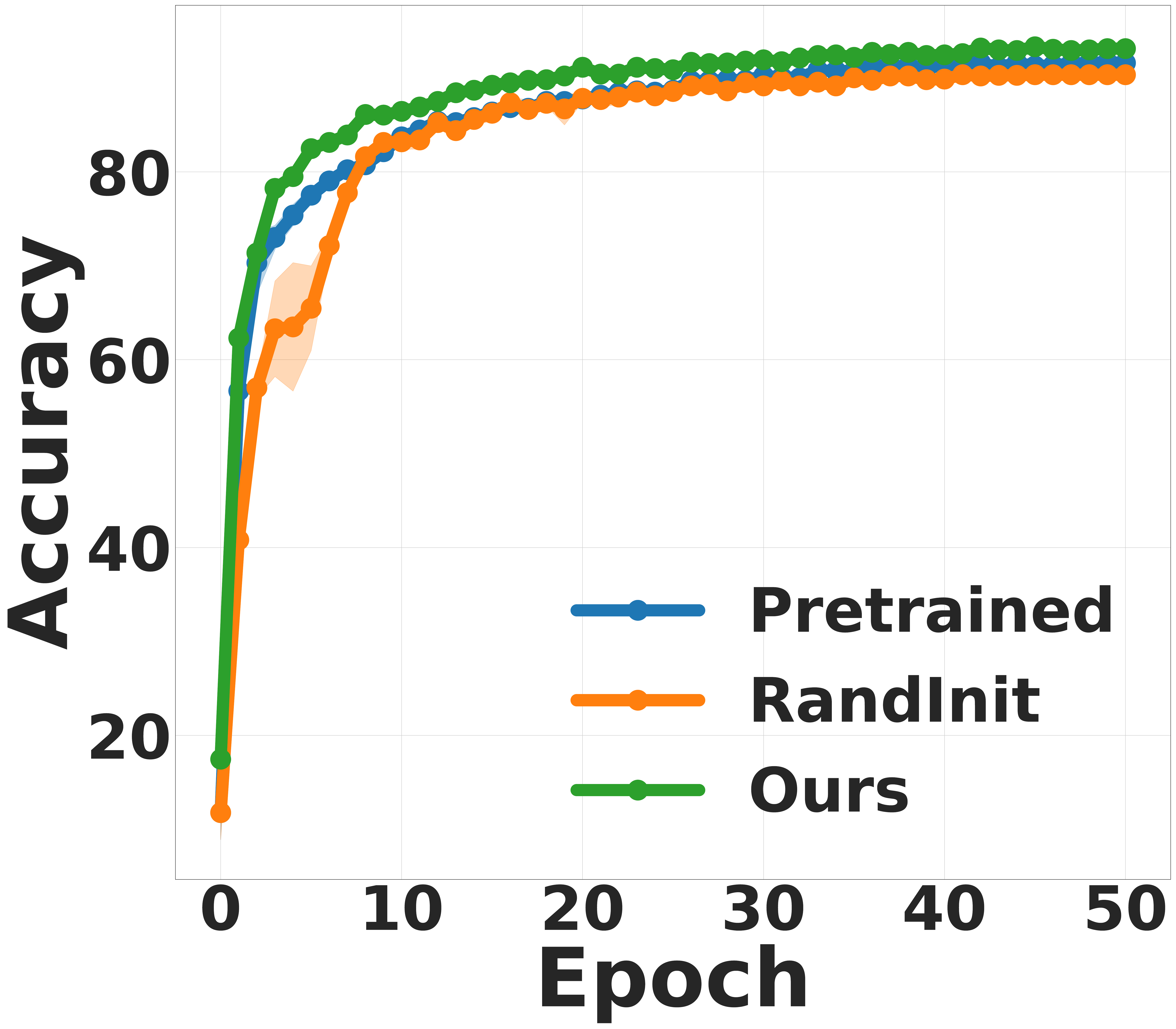}
	\captionsetup{justification=centering,margin=0.5cm}
	\caption{Colorectal Histology}
    \end{subfigure}
    \caption{\small Average accuracy evolution of fine-tuning for 50 epochs with sampled weights for unseen datasets.}
    \label{fig:kaggle}
     \vspace{-0.15in}
\end{figure*}

%% file: figures/transfer.tex
\begin{wrapfigure}{r}{0.5\textwidth}
    \vspace{-0.15in}
% \begin{figure}
\centering
\includegraphics[width=1.0\linewidth]{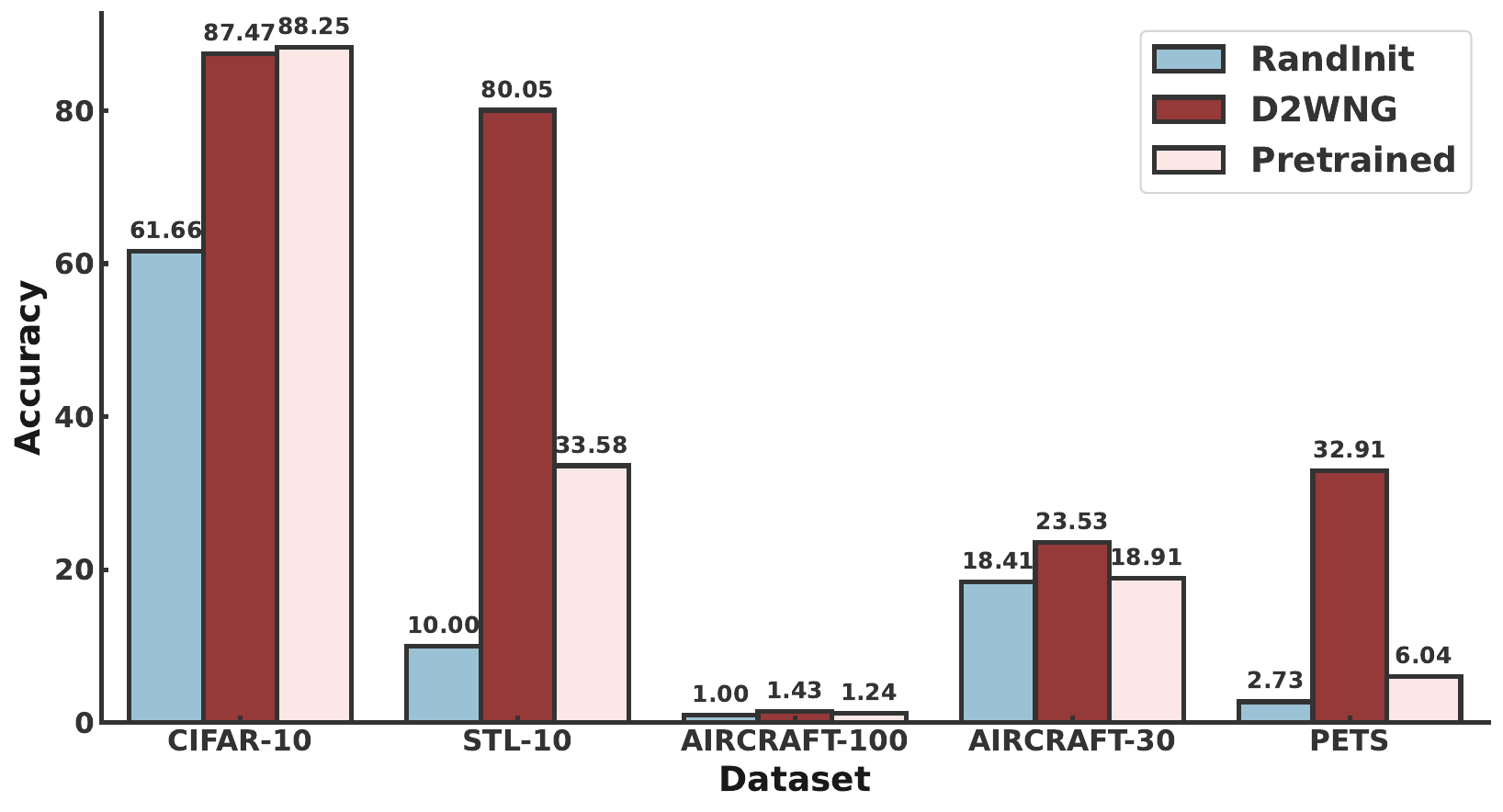}
    % \vspace{-0.05in}
    \caption[]{\small Comparison of accuracy for Pretrained, \ourmethod, and RandInit methods across CIFAR-10, STL-10, Aircraft100, Aircraft30, and Pets after 1 epoch of fine-tuning.}
    \vspace{-0.25in}
    \label{bartransf}
% \end{figure}
\end{wrapfigure}

%% file: tables/loraTable.tex
\begin{table*}[!ht]
\caption{Task Conditioned LoRA parameters Generation. Adaptations are performed on a Roberta-Base model denoted Rob-B.}
\label{lora}
    \centering
    % \vspace{-.3cm}
  \resizebox{1.0\textwidth}{!}{
    \begin{tabular}{lllllllllll}
   \toprule
        Method & \begin{tabular}[l]{@{}l@{}}Parameters\end{tabular}  & SST-2 (Acc) & MRPC (Acc.) & CoLA MCC.) & QNLI (Acc.) & RTE (Acc.) & STS-B (PCC.) & Avg. \\
        \midrule
        Rob-B & 125M  & \textbf{94.8} & 90.2 & 63.6 & \textbf{92.8} & 78.7 & 91.2 & 85.2&\\
        LoRA & 0.9M & 95.1$\pm$0.2 & 89.7$\pm$0.7 & 63.4$\pm$1.2 & 93.3$\pm$0.3 & 78.4$\pm$0.8 & 91.5$\pm$0.2 & 85.2  \\
        AdaLoRA & 0.9M & 94.5$\pm$0.2 & 88.7$\pm$0.5 & 62.0$\pm$0.6 & 93.1$\pm$0.2 & 81.7$\pm$0.6 & 90.5$\pm$0.2 & 85.0 \\
        DyLoRA & 0.9M & 94.3$\pm$ 0.5 & 89.5$\pm$0.5 & 61.1$\pm$0.6 & 92.2$\pm$0.1 & 78.7$\pm$0.7 & 91.1$\pm$0.2 & 84.5 \\ 
        FourierFT & 0.6M &  94.2$\pm$0.3 & 90.0 $\pm$ 0.8& 63.8$\pm$1.6 & 92.2$\pm$0.1 & 79.1$\pm$0.5 & 90.80 $\pm$ 0.2 & 85.0 \\ 
        D2NWG & 0.6M  & 94.3$\pm$0.1 \textcolor{blue}{+0.2} & \textbf{90.3}$\pm$0.5($\uparrow$\textcolor{blue}{0.3}) & \textbf{64.3}$\pm$1.2 ($\uparrow$\textcolor{blue}{0.5}) & 92.6$\pm$0.2($\uparrow$\textcolor{blue}{0.5}) & \textbf{79.6}$\pm$0.4($\uparrow$\textcolor{blue}{0.5} )& 91.0$\pm$0.3($\uparrow$\textcolor{blue}{0.0.2}) & 85.3($\uparrow$\textcolor{blue}{0.3}) \\

        \bottomrule
    \end{tabular}
}
\end{table*}

%% file: tables/llm_task_specific.tex
\begin{table}[t]
\centering
\caption{Exploration of optimal weight space of some instruct LLMs using a diffusion model. These are the average raw obtained with the lm-harness tool without the flash-attention. $\uparrow $ indicates the performance gain}
\label{tab:task-spec}
    \resizebox{0.9\textwidth}{!}{
    \renewcommand{\arraystretch}{1.1}
    \renewcommand{\tabcolsep}{12pt}
    \begin{tabular}{lccccc}
    \toprule
        Methods & Winogrande (5 shot) & Arc-Challenge (25 shot) & Hellaswag (25 shot) \\
        \midrule
        LLAMA-3.1-8B-Instruct &67.17 $\pm$ 0.01  &64.93 $\pm$0.01 &78.58 $\pm$ 0.00 \\
         \ourmethod & 67.61$\pm$ 0.02($\uparrow $\textcolor{blue}{0.44}) & 65.74$\pm$0.01($\uparrow $\textcolor{blue}{0.81}) & 78.86$\pm$ 0.02($\uparrow $\textcolor{blue}{0.28})\\
         \midrule

         Mistral-7b-Instruct &69.93 $\pm$ 0.01  &59.22 $\pm$0.01 &81.97 $\pm$ 0.00 \\
         \ourmethod & 70.80$\pm$ 0.02($\uparrow $\textcolor{blue}{0.80}) & 59.80$\pm$0.01($\uparrow $\textcolor{blue}{0.58}) & 82.04$\pm$ 0.00($\uparrow $\textcolor{blue}{0.07})\\
         \midrule
    
         LLAMA-3.2-1B-Instruct & 56.75$\pm$ 0.01  &40.96 $\pm$0.01 &61.67 $\pm$ 0.00 \\
         \ourmethod & 57.17 $\pm$ 0.01($\uparrow $\textcolor{blue}{0.42}) & 41.55 $\pm$ 0.01($\uparrow $\textcolor{blue}{0.59}) & 61.70$\pm$ 0.01($\uparrow $\textcolor{blue}{0.03})\\
         % \midrule
         \bottomrule
    \end{tabular}}

\end{table}

%% file: tables/llm_leaderboard.tex
\begin{table}[t]
\caption{Performance evaluation on unseen open llms leaderboard v2 benchmark. These results are produced by Huggingface after submission to open LLM leaderdoards. $\uparrow $ indicate performance improvement while $\downarrow$ indicate a performance decrease}
\label{leader_board}
\centering
\resizebox{\textwidth}{!}{
\renewcommand{\arraystretch}{1.0}
\renewcommand{\tabcolsep}{4pt}
\begin{tabular}{lccccccccc}
\toprule
    % \textbf{Method} & \textbf{ifeval (0 shot)} & \textbf{Bbh (3 shots)} & \textbf{Gpqa (0 shots)} & \textbf{MATH-hard (4)} & \textbf{Musr (0 shot)} & \textbf{MMLU-Pro (5 shots)} & \textbf{Avg} \\ 
    \textbf{Method} & \textbf{ifeval (0)} & \textbf{Bbh (3)} & \textbf{Gpqa (0)} & \textbf{MATH-hard (4)} & \textbf{Musr (0)} & \textbf{MMLU-Pro (5)} & \textbf{Avg}&Base Model& Fine-tuned \\ 
\midrule

% openai-community/gpt2  & 17.8  &  2.83 & \textbf{1.12} & 0.3 & \textbf{13.91}  & 1.84 & 6.3& na & Yes\\ 
% \ourmethod      & \textbf{19.16}($\uparrow $\textcolor{blue}{1.36}) &\textbf{2.85}($\uparrow $\textcolor{blue}{0.02}) & 1.01($\downarrow $\textcolor{blue}{0.11})  & \textbf{0.3}8($\uparrow $\textcolor{blue}{0.08}) & 12.68($\downarrow $\textcolor{blue}{1.23}) & \textbf{1.68}($\downarrow $\textcolor{blue}{0.16}) & 6.29($\gownarrow $\textcolor{blue}{0.01}) &Meta-openai-community/gpt2 & No \\ 
% \midrule
% \midrule
    Meta-Llama-3.2-1B-Instruct  & 56.78  & 8.74  & \textbf{3.36} & \textbf{2.96} & \textbf{2.97}  & 7.58 & 13.76& Meta-Llama-3.2-1B & Yes\\ 
\ourmethod      & \textbf{58.44}($\uparrow $\textcolor{blue}{1.66}) &\textbf{8.82}($\uparrow $\textcolor{blue}{0.08}) & 1.68($\downarrow $\textcolor{blue}{1.68})  & 6.04($\uparrow $\textcolor{blue}{3.08}) & 0.66($\downarrow $\textcolor{blue}{2.31}) & \textbf{9.09}($\uparrow $\textcolor{blue}{1.51}) & \textbf{14.12}($\uparrow $\textcolor{blue}{0.36}) &Meta-Llama-3.2-8B-Instruct & No \\ 
\midrule
    SauerkrautLM-8B-Instruct         & 80.17  & 31.00  & \textbf{5.37}  & 11.18 & 11.52 & \textbf{32.12} & 28.56 &Meta-Llama-3.1-8B-Instruct & Yes \\ 
    \ourmethod      & \textbf{80.33} \textcolor{blue}{+0.16}  & \textbf{31.10}($\uparrow $\textcolor{blue}{0.10} ) & 5.26($\downarrow $\textcolor{blue}{0.11})  & \textbf{11.56}($\uparrow $\textcolor{blue}{0.38}) & \textbf{11.52} & 32.07($\downarrow $\textcolor{blue}{0.05}) & \textbf{28.64} ($\uparrow $\textcolor{blue}{0.08}) & SauerkrautLM-8B-Instruct&No \\ 
% \midrule

\midrule
   Lexi-Uncensored-V2 & 77.92  & 29.69  & 4.36  & 16.92 & 7.77  & 30.90 & 27.93 & Meta-Llama-3.1-8B-Instruct & Yes\\ 
    Meta-Llama-3.1-8B-Instruct  & \textbf{78.56}  & 29.89  & 2.35  & \textbf{17.60} & 8.41  & 30.68 & 27.91& Meta-Llama-3.1-8B & Yes\\ 
\ourmethod      & 77.85($\downarrow $\textcolor{blue}{0.71}) &\textbf{30.39}($\uparrow $\textcolor{blue}{0.5}) & \textbf{4.47}($\uparrow $\textcolor{blue}{2.12})  & 17.52($\downarrow $\textcolor{blue}{0.08}) & \textbf{9.64}($\uparrow $\textcolor{blue}{1.23}) & \textbf{31.02}($\uparrow $\textcolor{blue}{0.34}) & \textbf{28.50}($\uparrow $\textcolor{blue}{0.59}) &Meta-Llama-3.1-8B-Instruct & No \\ 
\bottomrule
\end{tabular}}

\end{table}

%% file: sections/discussion.tex
In this work, we recast latent diffusion for dataset-conditioned neural network weight generation, enabling quick adaptation to novel datasets and efficient fine-tuning and transfer learning without training. Through extensive experiments on diverse datasets, our method generates high-quality weights for novel tasks and improves generalization. We extend parameter generation to large language models, demonstrating the scalability and versatility of our approach. Our method effectively encodes architectures with up to 1 billion parameters using a single GPU with less than 80GB, including task- or dataset-conditioned generation.

%% file: sections/appendix.tex
\section{Approach }\label{app:approach}
% We proposed a d
\textbf{Broader Impact} D2NWG addresses the resource-intensive nature of deep learning by proposing a method for efficient transfer learning. This has the potential to reduce the computational resources required for training neural networks, making it more accessible to a wider range of researchers and organizations.

\par \textbf{Limitation}
 In this work, we focus mainly on generalization across datasets. Additionally, while the diffusion model achieves impressive performance on image generation, there are still some challenges to efficiently recast it for weights generation including memory constraint, convergence challenges and considerations of symmetries in the weight spaces of different neural network architectures.

\subsection{Relationship Between Datasets and Trained Weights}\label{sec:former-theorem}
Gradient descent based optimization is the commonly used technique to generate optimal neural network weights through training by minimizing a loss function, ie. cross-entropy for classifiation tasks. The weights optimized with gradient descent thus contains some information about the training data. Therefore, understanding the correlation between the training dataset and the optimal weights is important for the generation of weights.  
During the optimization process with gradient descent the weights of each layer $i$ are updated as $w_i = w_{i-1}-\eta \nabla_{w_i} \mathcal{L}(w_1, w_2,\ldots, w_n)$, where $\nabla_{w_i} \mathcal{L}(w_1, w_2,\ldots, w_n)$ is input dependent. As an example, let's consider a two-layer feedforward neural network:
\begin{align*}
        x &: inputs & \quad & \quad\\
        l_1 &= W_1x +b_1 & h &= ReLU(l_1)\\
        h &= ReLU(l_1) & l_2 &= W_2 h+b_2\\
        \hat{y} &=softmax(l_2)& J &= CE(y, \hat{y})
    \label{eq1}
\end{align*}
Analyzing the weights' update below, we can observe that the optimal weights are noisy perturbation of the inputs feature maps and all together they contain information about the training either related to the raw input or the feature map at a given stage.
\begin{align*}
     \delta_1 & = \frac{\partial J}{\partial l_2} =(y-\hat{y})^T\\
    \delta_2 & = \frac{\partial J}{\partial l_1} =\delta_1 W_2 o sgn(h) \\
    W^{(i+1)}_1 &= W^{(i)}_1 -\eta \nabla_{w_1} \mathcal{L}(w_1, w_2,b_1, b_2)\\
    &=W^{(i)}_1 -\eta \delta_{2}^{T} x\\
    W^{(i+1)}_2 &= W^{(i)}_2 -\eta \nabla_{w_2} \mathcal{L}(w_1, w_2,b_1, b_2)\\
    &=W^{(i)}_2 -\eta \delta_{1}^{T} h^T
\end{align*}
%\vspace{-0.15in}
%\end{proof}

\subsection{Weights Vectorization}
For a neural network with $L$ layers, the process of vectorizing the weights and biases for both fully connected and convolutional layers is as follows:\
\begin{itemize}
    \item For the $\ell$'th  fully connected layer: 
  $W^{(l)} \in \mathbb{R}^{d_{l-1} \times d_l} \rightarrow \text{vec}(W^{(l)}) \in \mathbb{R}^{d_{l-1}.d_l}$ and $b^{(l)} \in \mathbb{R}^{d_l}$, the length of the vectorized weights for this layer, including the bias if it is not null, is given by \( d_{l-1}d_l + d_l \).
  \item For the $\ell$'th convolutional layer: 
  $W^{(l)} \in \mathbb{R}^{k_h . k_w.c_{in}.c_{out}}$ and $b^{(l)} \in \mathbb{R}^{c_{out}}$, the length of the vectorized weights for this layer, including the bias if it is not null, is \( k_h \cdot k_w \cdot c_{in} \cdot c_{out} + c_{out} \).
\end{itemize}
We then concatenate all the flattened weight and bias vectors  resulting in a vector $\theta$: $ \theta = \bigoplus_{l=1}^{L} \left( \text{vec}(W^{(l)}) \oplus b^{(l)} \right)$\
where $\text{vec}$ denotes the vectorization operation and $\oplus$ denotes concatenation.The concatenation operation keeps the ordering of weights in the network.

\subsection{Layer Selection Strategy}\label{app:layer_selection}
To manage the large number of parameters in LLM architectures, where not all layers are required to be tuned to improve the performance, we propose focusing on the most important layers. These layers are identified using the Marchenko-Pastur (MP) distribution, which serves as a filter to highlight relevant weights while discarding those resembling random noise. The MP law provides a benchmark for distinguishing structured weights from noise by comparing the empirical eigenvalue spectrum of weight matrices to the MP distribution.
D2NWG uses this \textit{spectrum method} \citep{hartford2024spectrumtargetedtrainingsignal} to learn the distribution of the most informative weights—those corresponding to eigenvalues that significantly exceed the MP upper bound. By focusing on these critical weights, D2NWG captures meaningful patterns in LLMs, leading to enhanced performance in transfer learning.

The spectrum method, grounded in random matrix theory, applies the Marchenko-Pastur (MP) distribution to different types of layers, treating them as rectangular random matrices. In transformer networks, functionally similar layers are grouped, such as  a set for all query layers in multi-head attention.
The method begins by computing the covariance matrix of each layer’s weight matrix, \( W \in \mathbb{R}^{m \times n} \), as \( \Sigma = \frac{W^T W}{n} \), followed by eigenvalue extraction. Singular value decomposition (SVD), \( W = U S V^T \), is used to efficiently compute these eigenvalues from the diagonal matrix \( S \), which contains the singular values. The resulting eigenvalues describe the variance captured by each principal component of the squared weight matrix and form what is known as the \textit{empirical spectrum}.
To analyze this spectrum, we compare it to the theoretical distribution of eigenvalues predicted by the Marchenko-Pastur (MP) distribution. This distribution $p(\lambda)$, in equation~\ref{marp}, characterizes the eigenvalue behavior of random covariance matrices as \( m, n \to \infty \), with a fixed aspect ratio \( q = \frac{m}{n} \) and variance \( \sigma^2 \).
% Such a comparison helps assess the structure and randomness in the learned weight matrices.
\begin{equation}
    p(\lambda)= \frac{1}{2 \pi\sigma^2 q\lambda}\sqrt{(\lambda_+-\lambda)(\lambda-\lambda_-)},
    \label{marp}
\end{equation}
where $\lambda \in \left[\lambda_+, \lambda_-\right]$, $\lambda_+=\sigma^2(1+\sqrt{q})^2$, and $\lambda_-=\sigma^2(1-\sqrt{q})^2$. From \ref{marp}, the correspoding  bounds for eigen values of $W$ are $\sqrt{\lambda}/\sqrt{n} \in \left[\varepsilon_+, \varepsilon_-\right]$, $\varepsilon_+=\frac{1}{\sqrt{n}}\sigma(1+\sqrt{q})$, and $\varepsilon_-=\frac{1}{\sqrt{n}}\sigma(1-\sqrt{q})$.

\textbf{Interpretation}: The Marchenko-Pastur (MP) distribution provides insight into the underlying structure of data or layer in our case:
\begin{itemize}
    \item \textit{Eigenvalues within MP bounds}: Likely represent noise, with their corresponding principal components carrying little meaningful information, indicating the layer's lower importance.
    \item \textit{Eigenvalues larger than the upper MP bound $\lambda_+$}: Capture more variance than noise, suggesting the presence of true signals or patterns in the data.
    \item \textit{Eigenvalues smaller than the lower MP bound $\lambda_-$}: May indicate compression or degeneration in the data structure.
\end{itemize}

Significant deviations, particularly large eigenvalues, indicate meaningful components that capture more variance than random noise, aiding in the identification of important features or signals. This insight is used to compute the signal-to-noise ratio (SNR), where eigenvalues below the upper bound are considered noise. The SNR is calculated as follows:
\begin{equation}
SNR = \frac{\sum_{k \, | \, |\sigma_k| \geq \varepsilon} \sigma_k}{\sum_{n \, | \, |\sigma_n| < \varepsilon} \sigma_n}.
\label{snr}
\end{equation}

\subsection{Learning the Distribution of LLM Weights}
Our method for LLM weight generation employs a layer-wise chunking mechanism that facilitates both layer-wise and chunk-wise sampling. Each layer is divided into independent chunks to form the training data, and are then encoded with the VAE. During the diffusion process, an index is assigned to each chunk, and the model is trained using class-conditioned diffusion, where chunk indices serve as class labels. At sampling time, the chunk indices corresponding to each layer are grouped into clusters associated with that layer. These clusters are then used to sample new sets of chunks, which are concatenated to reconstruct the sampled weights for each layer.

After selecting the top 25\% of the layers, we applied chunking with a size of 2,097,152 for LLaMA 3.2-1B and 4,194,304 for other models. We then performed sequential refinement using Algorithm \ref{algo_llm}. Unlike in vision tasks, LLM models are conditioned on chunk indices. Here, we refer to neural network operations such as dense layers and layer normalization as \textit{layers}. The spectrum method provides an ordered set of these layers (q, k, v, o, mlp\_up, mlp\_down, mlp\_gate). For architectures like Llama 3.1-8B and Mistral, we only learn the distribution of the top 8 each of these layers, excluding layer normalization. These layers are further divided into two groups: the top 4 and the second top 4, for which we build separate models to learn their distributions. As for the normalization layers, we learn the distribution across all of them. The maximum generated parameters is $\approx$ 872M.
\input{algorithms/algo_llm}

\subsection{Details of Modelzoo Generation}\label{detail:gen}
% The code source will be released on Github \url{https://github.com/sorobedio/DNNWG}
\subsection{Modelzoo and Pretrained Datasets}\label{zoos}
\par \textbf{Model zoo} We use the pretrained datasets from \citet{schurholtModelZoosDataset2022} as structured in \cite{schurholtHyperRepresentationsGenerativeModels2022}. This dataset consists of 4 different datasets with 5000 pretrained weights per architectures and datasets. The details of the architecture used to generate the pretrained weights are available in \citet{schurholtModelZoosDataset2022}. 
\par \textbf{KaggleZoo}
This modelzoo is generated using the dataset provided by~\citet{Jeong2021TaskAdaptiveNN}.
To efficiently generate the pretrained weights, we first compute the features of each image then use a MLP with two layers with input size 512, hidden size 256 and leaky ReLU activation functions. We train the MLP on clip features as it allows us to quickly generate high performing weights. For each datasets we used the last 10 checkpoints which results in 1400 pretrained weights for training.

\par \textbf{ImageNet zoo}
To generate the pretrained modelzoo on ImageNet, we sample 1000, 5000, 10000 and 20000 subsets with 10 classes each with 100 images per class in the training set and 50 per class in the test set. For the 1000 and 5000 subsets we used the same MLP architecture as the KaggleZoo. For the 10000 subset, we reduce the hidden dimension to 128 and, for the 20000 subset we use a single linear probing layer. On the other datasets linear probing  shows similar generalization performance as the two-layer MLP. We use Adam optimizer with a learning rate of $1e-3$ and all models are trained for 30 epochs.
\par \textbf{Zoo for Few-shot learning}: The few-shot learning pretrained zoo is generated by fine-tuning the classifier head for 10 epochs on each of the 50,000 subsets.
\par \textbf{LLMs zoo}: We collected the pretrained LLM model from their original HugginFace repositories with no further pertaining on specific tasks or datasets.

\par \textbf{Meta-album datasets}: We split the meta-album dataset into a training set (70\%) and a test set (30\%). Next, we trained the MobileNetV3 OFA subnet with parameters \( d = 2 \), \( k = 3 \), and \( e = 3 \) for 100 epochs. Checkpoints from the last 20 epochs were collected as training data. A detailed breakdown of the dataset can be found in Table \ref{tab:dataset_table}.

\subsection{Details of the Proposed Model}\label{detail}
We build our dataset conditioned weight generation model using latent diffusion \citep{Rombach2021HighResolutionIS}.
\par \textbf{AutoEncoder}: We use the same VAE modules of latent diffusion and use the same architecture for all experiments except adaptation of the inputs and output dimensions. We insert a linear layer before the first layer of the encoder such that we can reshape its output to a representation for the convolution layers. Similarly, a linear layer is placed at the last layer of the decoder adapting the output to the vectorized weights representations. For the VAE loss function we removed the discriminator in the original latent diffusion VAE loss function.

\input{tables/config_table}

\par \textbf{Diffusion Model}: We utilize same UNet architecture as in latent diffusion with the same training procedure.

\par \textbf{Dataset Encoding Mechanisms}
We investigated three different mechanisms of dataset encoding. Firstly, we use Set Transformer~\citep{lee2019set} which can be difficult to train when optimized together with the diffusion using the weights encoder from the VAE and the Set Transformer.

In addition to the Set Transformer, we explored a two-layer MLP model as the dataset encoder. The first layer is a dynamic linear layer with a maximum input feature size set to \(n_{\text{max}} \cdot c_{\text{max}}\), where \(n_{\text{max}}\) is the maximum number of images per class and \(c_{\text{max}}\) is the maximum number of classes among all subsets of the pretrained datasets. The shape of the image features in each dataset obtained with the CLIP image encoder is \( x \in \mathbf{R}^{c \times n \times d} \), where \( d \) is the feature dimension for each corresponding pretrained weight vector. While the Set Transformer-based encoder uses these inputs directly, the MLP encoder reshapes each input from \( x \in \mathbf{R}^{c \times n \times d} \) to \( x \in \mathbf{R}^{d \times (n \cdot d)} \) and then applies the dynamic linear layer.  If a dataset has more classes or samples than \( c_{\text{max}} \) and \( n_{\text{max}} \) respectively, we only consider the first \( c_{\text{max}} \) classes and \( n_{\text{max}} \) samples per class. If the dataset has fewer classes or samples, we adjust the dynamic linear layer dimensions accordingly. The output of the dynamic linear layer is \( z \in \mathbf{R}^{d \times h} \), where \( h \) is an arbitrarily chosen number greater than zero. We then reshape \( z \) from \( \mathbf{R}^{d \times h} \) to \( \mathbf{R}^{1 \times (h \cdot d)} \) (with \( h \cdot d \) fixed) and apply the final linear layer to obtain the desired output. This model can be jointly optimized with the diffusion model while achieving good performance.

\par \textbf{Dataset Encoding with Set Transformer} We use the Set Transformer for dataset encoding, pretrained as described in \cite{lee2021rapid}. The approach involves using the frozen Set Transformer and adding a single linear layer to adapt its output to our specific problem, utilizing it as the dataset encoder. This method reduces the computational cost of training the Set Transformer and enables joint optimization of the dataset encoder and the diffusion model. The results of these data set encoding schemes are presented in Table~\ref{tab:dataencoder} for the Hyperzoo dataset.
\input{algorithms/algo}

\section{Training Details}\label{train_detail}
In this section, we describe the training steps used to train our method.
\begin{itemize}
    \item \textbf{Pretrained Zoo Generation:} For classifier head adaptation, we first compute the features for all datasets. Then, we train the classifier head to generate the pretrained zoo. 
    
    \item \textbf{VAE Training:} We train the VAE to encode the pretrained weights following Equation~\ref{eqn:vae}. Additionally, a pretrained performance predictor can be used to predict the performance of the reconstructed weights and guide the VAE training as described in Equation \ref{eqreg}.
    
    \item \textbf{Dataset Alignment:} If using dataset alignment, we pretrain the Set Transformer to align the pretrained weights' latent representations. This is done using the frozen encoder of the VAE and the dataset embeddings. The inputs to the Set Transformer are image features, with five image features per class.
    
    \item \textbf{Diffusion Process Training:} We train the diffusion model while keeping the Set Transformer and the VAE models frozen. If an MLP is used for dataset encoding, we jointly optimize the diffusion process with the MLP dataset encoder.
\end{itemize}

Although the dataset encoder can be optimized together with diffusion model, we train them separately to speed up the training process and reduce memory requirements. The VAE and the dataset encoder are trained using the Adam optimizer with a learning rate of $1e-4$. The diffusion model in each experiment is trained with a linear scheduler, a base learning rate of 1e-4, and the AdamW optimizer~\citep{Rombach2021HighResolutionIS}. During the training process of the diffusion model, the output of the dataset encoder is concatenated with the latent representation of the input weights, forming the input to the UNet model. Additionally, we investigate joint training of the diffusion process in the ablation study and Appendix~\ref{abl1} and ~\ref{detail}. Further details can be found in Table~\ref{app.config}.

\input{figures/datasetencoder}

\subsection{Predictor Training}
To improve the reconstruction and sampling efficiency, we trained an accuracy predictor $g$ from pretrained weights $w$ then use the frozen predictor during the training of the VAE as a regularizer as shown below:
\begin{equation}
    \underset{\theta, \sigma}{\min}  \frac{w-f_{\theta}(w)}{\sigma^2}+\log \sigma^2 + ||g(w)-g(f_{\theta}(w))||^2
    ,\label{eqreg}
\end{equation}
where $g(w)$ is the embedding of the original input and $g(f_{\theta}(w))$ is the predictor embedding of the reconstructed weights. The predictor can be either dataset-conditioned or unconditioned. In general we found that dataset-conditioned predictor works only well for large number of samples per dataset. After the AutoEncoder is trained, we train the dataset-conditioned module which requires a dataset encoder.
\input{algorithms/algo2}

\section{Ablation Study}\label{app:ablation}

\subsection{Can the proposed method handle multiple architectures?}
This section provides a simple way to handle the case where the pretrained zoo contains multiple architectures per task or dataset.
Since the number of architecture and dataset are predefined, it is possible to build a set of unique index for each combination of dataset-architecture pairs.  An alternative will be to encode the graph representation of the architectures then used that as conditioning. In this ablation study we use the simple class indexing approach to demonstrate the versatility of our method.
We use CIFAR10 and CIFAR100 as the dataset and as target architectures we utilze a ResNet44 trained on CIFAR-100 with  667,188 parameters and a
ResNet44 trained on CIFAR-10 with 661,338 parameters and finally, a  MobileNetV2 trained on  CIFAR-10 with 700,490 parameters. All models were zero-padded to 700,490 parameters, combined into a unified dataset, and trained without chunking. The results in Table \ref{mix_archs} demonstrate that the proposed method is capable of simultaneously learning the distributions of diverse architectures trained on diverse datasets.
\input{tables/mixtureofarchitectures}

\subsection{Transferability}

We conducted a set of experiments using ResNet32 pretrained on CIFAR-10 and CIFAR-100, as the base model to explore the transferability of learned weights. Our objective is to model the distribution of the combined pretrained weights conditioned on their respective datasets.

We then evaluated these conditionally sampled weights on CIFAR-10 across different architectures, including ResNet20, ResNet44, ResNet56, and ResNet32. Inspired by GHN3’s weight tiling approach for larger networks, we adopted the following strategy: instead of tiling, we sampled multiple weights from the base model and concatenated them to match the dimensionality of the target architectures.

This concatenation was carefully structured to align each layer type between the base and target networks. The resulting vector is then used as initialization for the unseen architectures. Figure~\ref{bar12} summarizes our results where "RandInit" refers to random initialization, and "Unseen" refers to architectures where our method was applied to pretrained weights without fine-tuning. Pre-trained models are referenced for comparison, and all models were sourced from publicly available models. From Figure~\ref{bar12}, it can be seen that our method achieved higher performance improvement at initialization on the unseen ResNet architectures.
\input{figures/cross-arch_1}

% In this section, we investigate the importance and limitation of the different modules used to build our method.

\subsection{Effect of Modelzoo Size Generalization}\label{abl3}
Here we investigates the impact of increasing the number of pretrained datasets on performance with experiments that use model zoos of sizes 5000, 10,000, and 20,000, derived from ImageNet subsets. Unseen target datasets CIFAR-10 and STL-10 are used. Sampling 50 weights, the average performance of the top 5 performing weights is shown in Figure~\ref{barmore}.
\par \textbf{Results:} On CIFAR-10 and STL-10, we obtain accuracies of $39.60\pm1.31\%$ and $44.66\pm0.55\%$ for 5000 subsets, $42.15\pm2.12$ and $64.83\pm2.83\%$ for 10000 subsets, and $52.64\pm3.12\%$ and, $80.49\pm1.77\%$ for 20000 subsets. 
The maximum accuracies with random initialization are $12.11\%$ and $17.12\%$ on CIFAR-10 and STL-10 without fine-tuning. This experiment demonstrated that increasing the number of datasets enhances the generalizability of the proposed method.
\input{figures/barplot1}

\subsection{Sampling without Latent Representation}\label{abl2}
\input{figures/barplot2}
This section explores a model variant that directly learns the diffusion model on weights, bypassing the AutoEncoder stage, and compares it to the standard approach. Both variants are trained on 1000 subsets of ImageNet, and evaluated in in-distribution sampling setting on three randomly selected subsets from the 1000 subsets. The results, presented in Figure~\ref{bar2}, indicate that learning the distribution of pretrained weights in the latent space is notably successful in generating high-performing weights. The failure of the DDPM process on raw pretrained weights may stem from their higher model capacity requirement.

\subsection{CLIP-based Dataset Encoding}\label{abl1}

In this section, the comparison between the CLIP-based dataset encoding scheme trained at an intermediate stage and the Set Transformer encoder jointly trained with the diffusion process is explored. Experiments are conducted on 140 Kaggle datasets and their respective model zoos.
The results depicted in Figure \ref{bar1} indicate that both methods achieve similar results for small numbers of datasets during the in-distribution sampling. However, as the number of datasets increases, the Set Transformer jointly trained with the diffusion approach faces challenges in convergence and requires more computational resources, as demonstrated in Figure \ref{bar1}.

\subsection{Unconditional Sampling}
We conduct the experiment using ResNet18 pretrained on CIFA-100 and CIFAR-10.
For all datasets, the weight vector length is 2048 and we compare with pdiff~\citep{wang2024neural}. While pdiff requires a separate model for each dataset, our method combines the pretrained weights into a single dataset and conditionally learns their distribution. The sample size for each dataset in our method is 200, with a combined total of 400 parameters. The results are provided in Table~\ref{pdiff} for 100 sampled weights. Two separate models for are trained for pdiff, CIFA10-pdiff and CIFAR100-pdiff while our method consists of a single model trained once for both datasets.  It can be seen that our method outperformance the baseline~\citep{wang2024neural} in Table~\ref{pdiff}.
\input{tables/d2wngvspdiff}

\subsection{Coupling with an Accuracy Predictor}
This section reports the extended results of Table \ref{wtab} in which we compared our method in-distribution and out-of distribution with and without accuracy predictor. 
\par \textbf{Results.}: The full results of Table~\ref{wtab} are reported in Table~\ref{wtab22}. Using an accuracy predictor enable easily selecting highly performing when sampling in-distribution. However, in our case the accuracy predictor struggles to generalize well for unseen dataset as shown in Table~\ref{wtab22}

\subsection{Sampled Weights Analysis}
In this section, we analyze the characteristics of the sampled weights and compare them to the pre-trained ones based on experiments with the model zoo and a model pre-trained on a subset of ImageNet. The proposed method samples weights with a large variance, as shown in Figure \ref{sim}, providing a broad range of initialization choices, from weights with low initial performance to those with higher initial performance.

\input{tables/metaalbum}

\input{tables/text_generator}

\input{tables/glues_tasks_descruiption}

\subsection{Evalutaion on Large Datasets}
We investigate how our method perform for combined large and small dataset as well for mixed architectures. For this experiment we collect the pretrained weights from PyTorch hub with one checkpoints per datasets(CIFAR-10, CIFAR-100, and ImageNet-1k). After conditionally learning the combined weights distribution, we sampled 10 weights for each datasets and report the average accuracy of the top-3 datasets in Table~\ref{table:mixed}. As shown in Table~\ref{table:mixed}, \ourmethod consistently produced high performing weights for each dataset from a single pretrained checkpoint.

\input{tables/mixed_arch}

\subsection{Generating the Full Weights for ResNet18}\label{app:full}
We investigate how our method performs when used to generate the full parameters for a ResNet18 model pretrained on MNIST, CIFAR-10, and CIFAR-100. In total, we use 100 pretrained weights per dataset and conditionally learn their distribution. The modelzoo generation follows the same setting as~\citet{wang2024neural}. Table~\ref{tab:res18} demonstrates the effectiveness of our method for generating the entire weights of a network.
\input{tables/zerishot_res18_in_dist}

\subsection{Generating Weights for MobileNetV3}
So far, our focus has been on model zoos populated by relatively simple classifier heads. In this section, we evaluate our method using MobileNetV3, a subnetwork sampled from OFA~\citep{cai2020once}, consisting of 2.8 million parameters fine-tuned on CIFAR-10, STL-10, SVHN and MNIST for 15 epochs. We collect the last 10 checkpoints per dataset and utilize our method to learn the distribution of pretrained weights.
Furthermore, we combine the pretrained weights of MNIST and CIFAR-10, learn their distribution, and then evaluate our method on SVHN and STL-10. Subsequently, we reverse this process by combining the pretrained weights of SVHN and STL-10, and evaluate our method on MNIST and CIFAR-10.

\par As shown in Table~\ref{tab:mbvv3} our method enhances the performance of the pretrained model. Furthermore, we note that learning the full model weights does not compromise performance. Although learning the distribution of the classifier head is computationally efficient, it can result in lower performance.

\subsection{Generating Weights for Vision Transformers}
Our method shows the ability to learn the distribution of all parameters within a vision transformer, including convolutional and linear layers. We present in-distribution evaluation results in plot Figure~\ref{barv1}, highlighting the learning of combined weight distributions conditioned on individual datasets. The model zoo for ViTs is collected based on models proposed by~\citet{Gani_2022_BMVC}.

\section{Application to Large Language Model (LLM) Output Layer Generation}\label{llm2}
\par\textbf{Phi-3-MINI-4K-Instruct:}We conduct experiments on the Microsoft Phi-3-MINI-4K-Instruct model to demonstrate the scalability of our method for generating output layers in large language models (LLMs). The model’s 98.5 million-parameter output layer was split into 96 chunks, each of size 1,026,048, and used as training data for a Variational Autoencoder (VAE) with an embedding size of 1,024. Lacking access to original training data, we used a class-conditional diffusion process, with chunk embeddings as conditioning data. Post-training, conditioned chunks were sampled and concatenated to reconstruct the original output vector. We evaluate our method using the Open-LLM LeadearBoard-1. 
As shown in Table~\ref{phi3}, our approach effectively scales to the LLMs head generation demonstrating adaptability across diverse domains with minimal adjustments to conditioning data.

\input{tables/phi3}
\par\textbf{GPT2:} In this experiment, we show that our method can learn the distribution of any layer in an LLM by modeling the full distribution of GPT-2 small (164M parameters). We use a chunk size of 1,523,712 and, unlike Llama architectures, concatenated all vectorized layer weights before chunking them uniformly. Table~\ref{gpt2} highlights the method's effectiveness on the Open LM-Leaderboard benchmark. While it did not outperform the base model overall, it significantly improved performance on certain tasks and maintained average accuracy comparable to the pretrained model.
\input{tables/gtp2}

\subsection{Fast Convergence Performance Evaluation}\label{shrol}
In this section we report supplementary results for experiment on tiny model zoo dataset. The pretrained weights used here are from epochs 21 to 25 for each dataset where 70\% of the resulting modelzoo is used for training and 15\% for validation and testing respectively. The number of pretrained weights in the modelzoos are 3500 for MNIST, CIFAR-10, and STL-10, and 2864 for SVHN. The flattened network weights' length is 2864 for CIFAR-10 and STL-10 and, 2464 for MNIST and SVHN. We pad all the weights with zero to 2864.

\input{figures/hyperzooplots}
\input{tables/dataset_transfer}

\subsection{Sampling Weights for Unseen Datasets}\label{sec-unseen}
\textbf{Task:} We evaluate the transferability of the models on unseen datasets. We create disjoint modelzoos by combining MNIST and CIFAR-10 into a single modelzoo and combining the SVHN and STL-10 modelzoos. When we train on the MNIST plus CIFAR-10 modelzoos, we test on the SVHN and STL-10 modelzoos and vice-versa.

\textbf{Results:} As shown in Table~\ref{wtab}, D2NWG is able to sample weights with higher accuracy on unseen datasets as well as for in distribution. Through these experiments our method does not only outperform the baseline it also demonstrates promising results for dataset-conditioned sampling for unseen datasets.

\input{tables/appendixtable2}
\input{tables/appendixtable4}
% Table~\ref{tab22} reports the detail results presented in Table~\ref{tab2}.
\input{tables/appendixtable3}
\input{figures/vit_bar}

\input{figures/similarity}

\section{Miscellanea}\label{recap}
% This section provide a details summary of all experiments.
\begin{table}[h]
\centering
\caption{Model components and their configuration modes for llma3.2.1B}
\begin{tabular}{lllll}
\toprule
\textbf{ID} & \textbf{Name}            & \textbf{Type}              & \textbf{Params} & \textbf{Mode} \\ 
\midrule
0           & Model                    & DiffusionWrapper           & 102 M           & Train         \\ 
1           & Model Ema               & LitEma                    & 0               & Train        \\ 
2           & First stage Model       & VAENoDiscModel            & 553 M           & Eval          \\ 
3           & Cond Stage Model        & IdentityCondStage         & 0               & Eval          \\ 
\bottomrule
\end{tabular}
\label{tab:model_configs}
\end{table}

% \subsection{Parameter counts}
In Table~\ref{tab:model_configs} we present the parameter count for the model used to learn the distribution of the 25\% of llama-3.2-1B transformer blocks. In Table~\ref{tab:param_label} we showcase the set of experiments and the corresponding number of parameters generated by \ourmethod. Although \ourmethod is capable of generating up to 1 billion parameters, all our experiments were limited to a maximum of 872 million, achieved using the Llama 3.1-8B model with 4 transformer layers, excluding layer normalization, for which we constructed a separate model. This parameter count makes \ourmethod the only method, to the best of our knowledge, capable of generating nearly a billion parameters, significantly enabling large architecture weights generation including GPT-2 and most existing image classification models in terms of parameter scale. 
%‘Train dataset’ refers to the number of unique pretraining datasets or tasks on which the model has been trained to generate the pretrained model zoos. ‘Dataset’ refers to the target task or dataset on which the model is tested through conditional sampling. 
For non-LLM models, we utilize joint distribution learning, enabling task or dataset-conditioned sampling. For example, CIFAR-10 and ImageNet are considered two separate datasets, while SST-2 and CoLA in the GLUE benchmark are treated as two distinct tasks, regardless of differences in the number of classes or subtasks within each dataset or task. 
%\textit{Min cls} and \textit{Max cls} denotes the minimum number of classes respectively the maximum number of classes. 
Table~\ref{tab:param_label} highlights that the proposed method supports text and image conditioning, as well as layer- or chunk-wise conditional sampling. \ourmethod is one of the first weight generation methods to produce over 800 million parameters in a single instance without tiling. Additionally, it is among the first to effectively explore weight generation across various domains, learning the distribution of combined models pretrained on diverse tasks or datasets.

\begin{table}
    \centering
        \caption{Summary of Experiments for Figures and Tables presented. Min \#cls and Max \#cls correspond to the minimum and maximum number of classes respectively. }
    \begin{tabular}{lccccccc}
    \toprule
        Object &\# Datasets & Min \#cls  & Max \#cls &\#Params   & Trainset Size & Conditioning \\
        \midrule
         Table~\ref{table_meta} & 10  & 1 & 5 & 2565/8005 & 50k & Dataset \\
         Table~\ref{tab:swin} & 5 & 10 & 100 & 128100 & 20k & Dataset   \\
         Table~\ref{tab:augmet} & 30 & 19 & 706 &3 M  & 30  &  Dataset  \\
         Table~\ref{tab:hyperzoo} &4  & 10 &10  & 10853 &4 & Dataset   \\
         Table~\ref{lora} & 6 & 2 & 3 & 0.6M & 6 & Text Description  \\   
         Table~\ref{tab:task-spec} & NA & NA & NA &872M  & NA &  Chunk Indices \\
         Table~\ref{leader_board}& NA & NA & NA &872M  & NA & Chunk Indices \\   
         Table~\ref{mix_archs}& 2 & 10 &  100& 0.7M & 2 & Dataset   \\
         Table~\ref{pdiff}  & 2 & 10 & 100 & 2048 & 2 & Dataset   \\     
        Table~\ref{table:mixed}  & 3 & 10 & 1000 & 1.4M & 3 & Dataset   \\
        Table~\ref{tab:res18} &  3& 10 &  100& 11M & 2 & Dataset  \\          
        Table~\ref{tab:res18} & 4 & 10 &  10& 2.8M& 4 & Dataset  \\
        Table~\ref{phi3} & NA & NA &  NA& 96M& NA & Chunk Indices  \\        
        Table~\ref{gpt2} & NA & NA &  NA& 164M& NA & Chunk Indices  \\
         Figure~\ref{fig:kaggle} & 10 &  2 & 1566 & 136468 & 140 &  Dataset  \\
        Figure~\ref{bar12} & 2 & 10 & 100 & 0.47M & 2 & Dataset   \\          
        Figure~\ref{barmore} & 2 & 10 & 10 & 5310 & 2 & Dataset   \\      
        Figure~\ref{bar1} & 2 & 10 & 10 & 5310 & 2 & Dataset    \\      
        Figure~\ref{barv1} & 5 & 10  & 200 & 2.8M & 5 & Dataset \\
    \bottomrule
    \end{tabular}
    \label{tab:param_label}
\end{table}

%% file: algorithms/algo_llm.tex
\begin{algorithm}
\caption{Sequential Weight Model Improvement}
\label{algo_llm}
\begin{algorithmic}[1]

\STATE \textbf{Input:} Initial weights $\Theta_{\text{init}} = \{\tilde{\theta}_1, \dots, \tilde{\theta}_L\}$, Hypernetwork $\mathcal{H}_i$ for each layer $i$, Validation dataset $\mathcal{D}_{\text{val}}$, $K$ candidates per layer
\STATE \textbf{Output:} Final weights $\Theta^* = \{\theta_1^*, \dots, \theta_L^*\}$

\STATE Initialize $\Theta^* = \Theta_{\text{init}}$
\STATE Compute initial validation accuracy: $\text{current\_accuracy} = \mathcal{A}(\Theta_{\text{init}}, \mathcal{D}_{\text{val}})$

\FOR{each layer $i = 1$ to $L$}
    \STATE Generate $K$ candidates $\{\theta_i^{(1)}, \dots, \theta_i^{(K)}\}$ using $\mathcal{H}_i$
    \FOR{each candidate $k = 1$ to $K$}
        \STATE Replace $\tilde{\theta}_i$ with $\theta_i^{(k)}$ in $\Theta^*$ to form $\Theta^{(k)}$
        \STATE Compute validation accuracy: $\mathcal{A}(\Theta^{(k)}, \mathcal{D}_{\text{val}})$
    \ENDFOR
    \STATE Choose $\theta_i^* = \arg\max_k \mathcal{A}(\Theta^{(k)}, \mathcal{D}_{\text{val}})$
    \IF{$\mathcal{A}(\Theta^{(k)}, \mathcal{D}_{\text{val}}) > \text{current\_accuracy}$}
        \STATE Update $\Theta^* = \Theta^{(k)}$
        \STATE Update $\text{current\_accuracy} = \mathcal{A}(\Theta^*, \mathcal{D}_{\text{val}})$
    \ELSE
        \STATE Retain $\tilde{\theta}_i$ in $\Theta^*$
    \ENDIF
\ENDFOR

\STATE \RETURN $\Theta^*$

\end{algorithmic}
\end{algorithm}

%% file: tables/config_table.tex
% \begin{table}[t]
\begin{wraptable}{r}{6cm}
\vspace{-0.15in}

% \vspace{-0.25in}
\caption{Models seting, $n$ and $c$ in the dataset configuration represent respectively the number of samples per class n=5 for training and c the total number of classes per dataset. The VAE and the diffusion models share similar configuration and architectures as \citep{Rombach2021HighResolutionIS}}\label{app.config}

\centering
\resizebox{0.4\columnwidth}{!}{
\begin{tabular}{lcc}\\
\toprule  
Parameters & Values\\
\midrule
Epochs & [50, 2000]\\
\midrule
\multicolumn{2}{c}{VAE}\\
\midrule
Optimizer &  Adam \\
Learning Rate & 1e-3  \\
Latent Dimiension & 1024\\
KL-Divergence Weight &  1e-6 \\
\midrule
\multicolumn{2}{c}{Dataset Encoder}\\
\midrule
Architecture &  Set Transformer \\
Input Dimension & $c\times n\times 512$(min)  \\
Output Dimension & 1024 (min)\\
Depth of Set Transformer &  2 \\
\midrule
\multicolumn{2}{c}{Diffusion}\\
\midrule
Optimizer& AdamW\\
Learning Rate & 1e-4\\
Scheduler & Linear \\
Time step & 1024\\
Network & Unet\\
UNet Input Size & $(c\times 32\times 32)$\\
\bottomrule
\end{tabular}
}
\vskip -0.10in
% \end{table} 
\end{wraptable}

%% file: algorithms/algo.tex
\begin{algorithm}[!t]
   \caption{Datasets Encoder Training}
   \label{alg1}
\begin{algorithmic}
   \STATE {\bfseries Input:} pretrained weights $x$, image features $y$, batch\_num $m$
   \STATE Instanciate $\mathcal{T}={\verb|Set Transformer|}$, Load pretrained Encoder ($\mathcal{E}$).
   \REPEAT
   \STATE Initialize $loss=0.0$
   \FOR{$i=1$ {\bfseries to} $m-1$}
   \STATE $x_i \sim x$, $\mathfrak{D_i} \sim \mathfrak{D}$
   \STATE $z_i = \text{Encoder}_{\text{VAE}}(x_i)$ %\mathcal{E}_{\theta}.encode(x_i).sample()$
   \STATE $z_{\mathfrak{D}_i} = \mathcal{T}(\mathfrak{D_i})$
   \STATE $loss = loss + \mathcal{L}_{CLIP}(z_i, z_{\mathfrak{D_i}})$ (Equation~\ref{eq2})
   \ENDFOR
   \STATE Update weights of $\mathcal{T}$
   \UNTIL{convergence}
\end{algorithmic}
\end{algorithm}

%% file: figures/datasetencoder.tex
\begin{figure}[t]
\begin{center}
\includegraphics[width=0.80\textwidth]{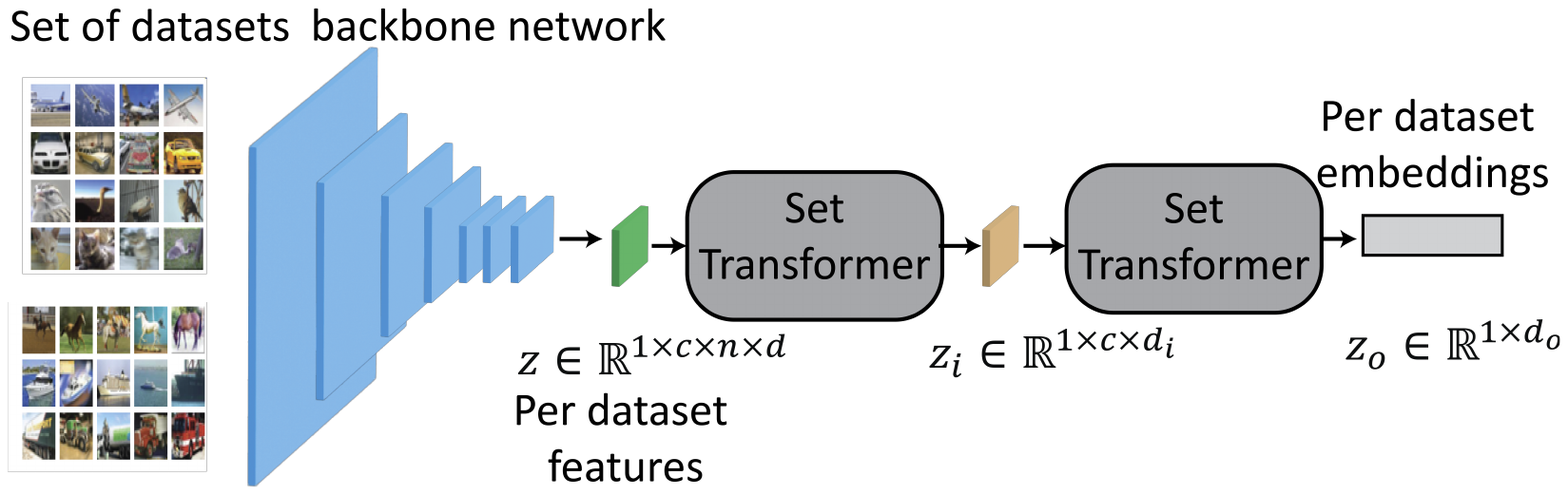}
\end{center}
\vspace{-0.05in}
\caption{Overview structure of the set-transformer-based dataset encoder. For each pretrained dataset we use $n=5$ images per class and the embedding dimension $d_0=1024$. }
\label{fig:setransf}
\end{figure}

%% file: algorithms/algo2.tex
\begin{algorithm}[tb]
   \caption{Predictor-Guided VAE}
   \label{alg2}
\begin{algorithmic}
   \STATE {\bfseries Input:} Pretrained weights $x$, accuracy $y$, batch\_num $m$
   \STATE Instantiate $f=\text{Set Transformer}$, and load pretrained predictor $g$).
   \REPEAT
   \STATE Initialize $loss=0.0$
   \FOR{$i=1$ {\bfseries to} $m-1$}
   \STATE $\Bar{x}=f_{\theta}(x)$, $\Bar{y}=g(\Bar{x})$ $\hat{y}=g(x)$
   % \STATE $\mathcal{L}=\frac{x-\Bar{x}}{\sigma}+\log \sigma + ||\hat{y}-\Bar{y}||^2$
   \STATE $\underset{\theta}{L} \frac{x-\Bar{x}}{\sigma^2}+\log \sigma^2 + ||\hat{y}-\Bar{y}||^2 $
   % \STATE $loss += \mathcal{L}_{clip}(z_i, z_q)$ (eq.\ref{eq2})
   \ENDFOR
   \STATE Update weights of $f$
   \UNTIL{Convergence}
\end{algorithmic}
\end{algorithm}

%% file: tables/mixtureofarchitectures.tex
\begin{table}[h]
\centering
\begin{tabular}{llll}
\toprule
Model            & ResNet44 (CIFAR-10) & ResNet44 (CIFAR-100) & MobileNetV2 (CIFAR-10) \\ 
\midrule
Pretrained         & 94.01                        & 71.63                         & 92.88                          \\ 
\ourmethod    & 94.10 $\pm$0.09                        & 71.64$\pm$0.02                            & 93.11$\pm$0.20                             \\ 
\bottomrule
\end{tabular}
\caption{Performance evaluation on mixed architectures.}
\label{mix_archs}
\end{table}

%% file: figures/cross-arch_1.tex
% \begin{wrapfigure}{r}{0.6\textwidth}
    \vspace{-0.15in}
\begin{figure}[h!]
\centering
\includegraphics[width=0.6\linewidth]{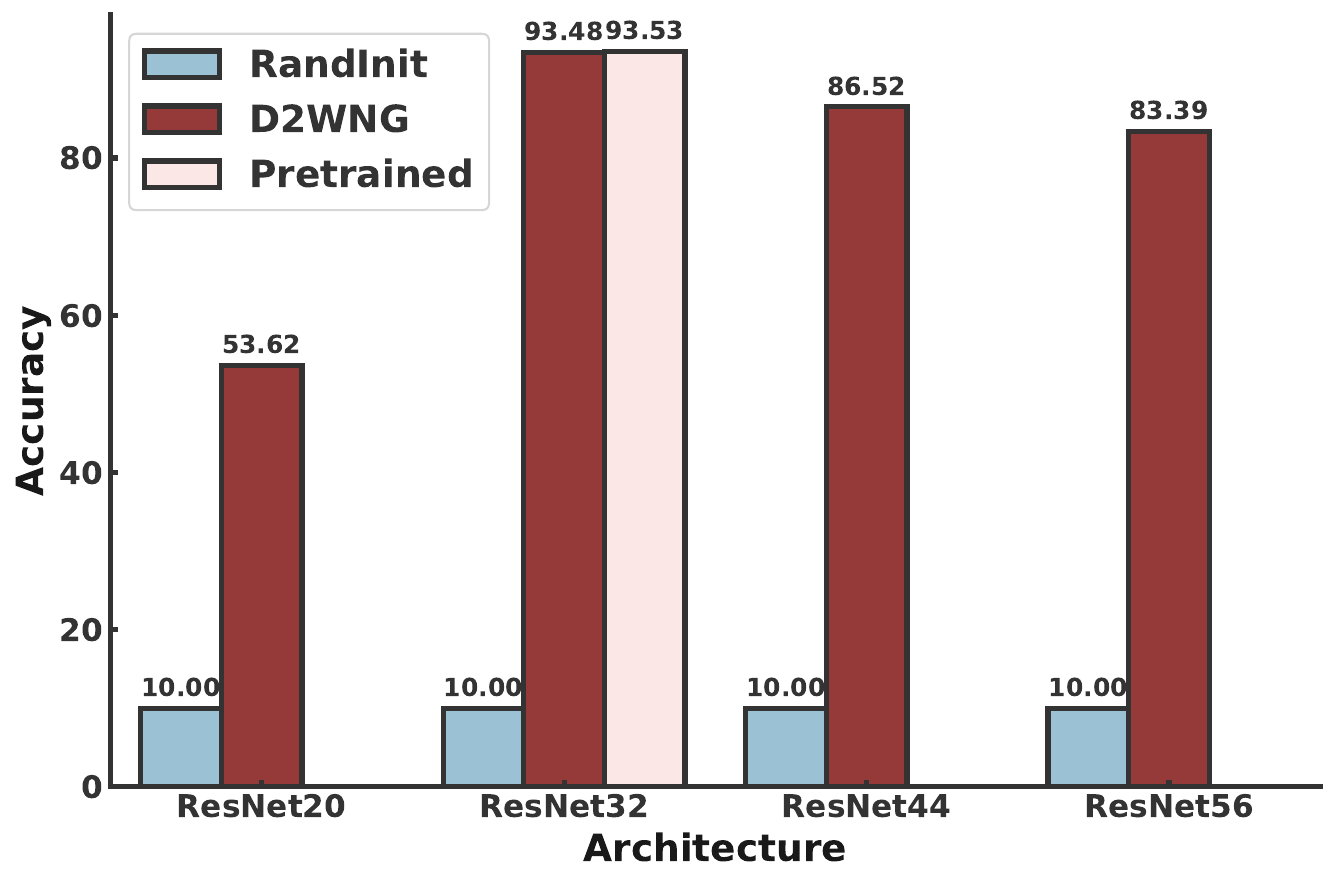}
    % \vspace{-0.05in}
    \caption[]{\small Performance evaluation with unseen architectures on CIFAR-10.}
    \vspace{-0.15in}
    \label{bar12}
\end{figure}
% \end{wrapfigure}

%% file: figures/barplot1.tex
% \begin{wrapfigure}{r}{0.4\textwidth}
    % \vspace{-0.05in}
\begin{figure}[h!]
\centering
    \begin{subfigure}{0.35\textwidth}
        \centering
        \includegraphics[width=\linewidth]{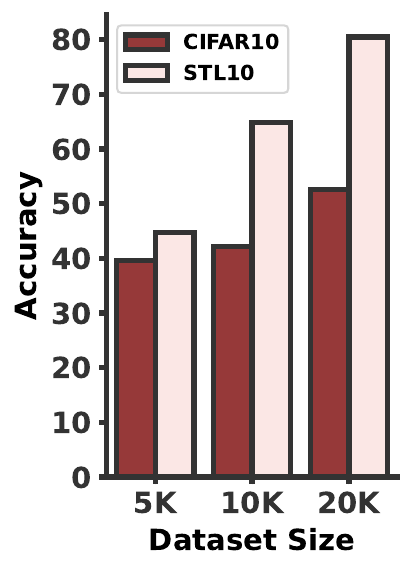}
        \vspace{-0.3in}
        \caption{}
        \label{barmore}
    \end{subfigure} 
    \begin{subfigure}{0.35\textwidth}
	\centering
	\includegraphics[width=\linewidth]{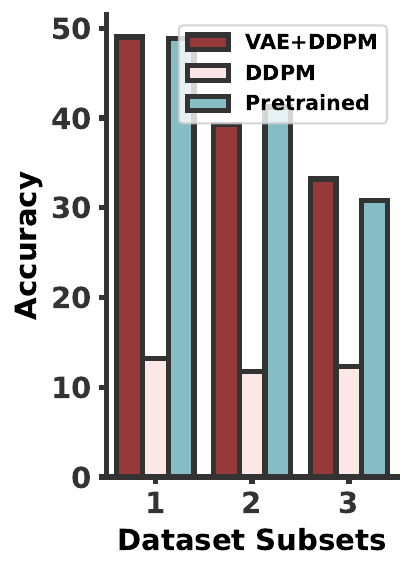}
	\captionsetup{justification=centering,margin=0.5cm}
	\vspace{-0.3in}
	\caption{}
	\label{bar2}
    \end{subfigure}
    %\vspace{-0.1in}
    \caption[]{\small (a) Effect of the number of pretrained datasets on sampling weights performance on unseen datasets. (b) Performance comparison on in-distribution sampling of methods with VAE+DDPM vs DDPM}
    \vspace{-0.15in}
    \label{bar:fig}
\end{figure}
% \end{wrapfigure}

%% file: figures/barplot2.tex
% \begin{wrapfigure}{r}{0.4\textwidth}
    \vspace{-0.15in}
\begin{figure}
\centering
\includegraphics[width=0.6\linewidth]{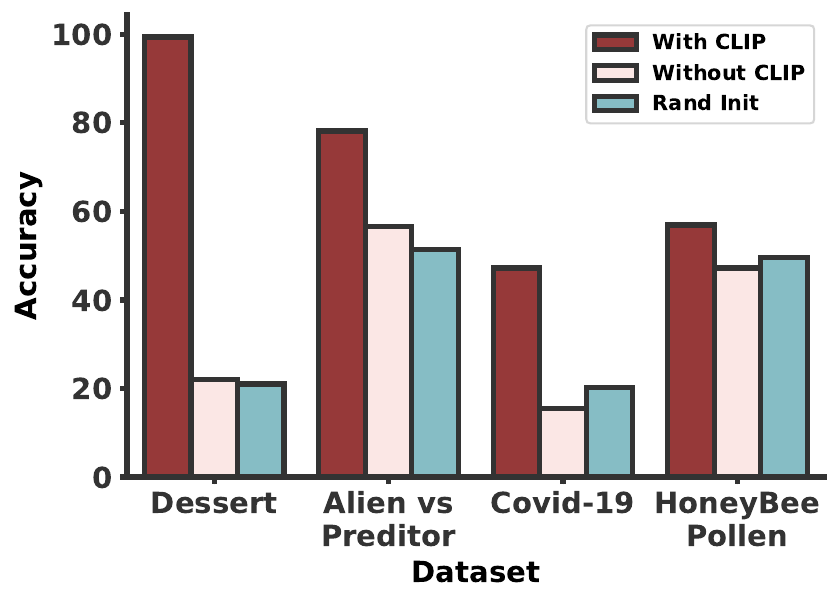}
    % \vspace{-0.05in}
    \caption[]{\small Performance comparison at initialization of method with jointly trained set-transformer (Without CLIP) and method clip-based dataset encoder.}
    \vspace{-0.25in}
    \label{bar1}
\end{figure}
% \end{wrapfigure}

%% file: tables/d2wngvspdiff.tex
\begin{table}[h!]
  \caption{Unconditional Sampling Evaluation against ~\citet{wang2024neural} on ResNet18.}
  \label{pdiff}
  \centering
  \resizebox{1.0\textwidth}{!}{
\begin{tabular}{lllll|lllll}
 \toprule
Dataset & \multicolumn{4}{l}{CIFAR-10}  & \multicolumn{4}{l}{CIFAR-100}   & Runtime              \\ 
\midrule
Method   & \multicolumn{1}{l}{Avg}     & \multicolumn{1}{l|}{Median} & \multicolumn{1}{l}{Max} & \begin{tabular}[l]{@{}l@{}}\#Epochs for\\VAE,DDPM\end{tabular} & \multicolumn{1}{l}{Avg} & \multicolumn{1}{l}{Median} & \multicolumn{1}{l}{Max} & \begin{tabular}[l]{@{}l@{}}\#Epochs for\\VAE,DDPM\end{tabular} &  \\ 
\midrule
pdiff    & \multicolumn{1}{l}{94.46} & \multicolumn{1}{l}{94.46}      & \multicolumn{1}{l}{\textbf{94.52}}    &  8999,47999  & \multicolumn{1}{l}{76.1028}    & \multicolumn{1}{l}{76.13}       & \multicolumn{1}{l}{76.21}    &  32999,38999  & $\approx 3h$      \\ 

D2NWG    & \multicolumn{1}{l}{94.46} & \multicolumn{1}{l}{\textbf{94.47}}       & \multicolumn{1}{l}{94.50}    &  100,200         & \multicolumn{1}{l}{\textbf{76.1796}}    & \multicolumn{1}{l}{\textbf{76.18}}       & \multicolumn{1}{l}{76.24}    &       100,200  & $\approx 1h30$ \\
    \bottomrule
\end{tabular}
}
\end{table}

%% file: tables/metaalbum.tex
\begin{table}[ht]
\caption{Details description of meta-album datasets}
\label{tab:dataset_table}
\centering
\begin{tabularx}{\textwidth}{XXX}
\toprule
\textbf{Domain} & \textbf{Original Dataset} & \textbf{\# Classes} \\ 
\midrule
Large Animals & Animals with Attributes, Dogs, Birds & 50, 120, 315 \\ 
Small Animals & Insects, Insects 2, Plankton & 117, 102, 102 \\ 
Plants & Fungi, PlantNet, Flowers & 25, 25, 102 \\ 
Plant Diseases & PlantDoc, Medicinal Leaf, Plant Village & 27, 26, 38 \\ 
Microscopy & Subcel. Human Protein, PanNuke, Bacteria & 21, 19, 33 \\ 
Remote Sensing & RSD, RSICB, RESISC & 43, 45, 45 \\ 
Vehicles & Boats, Airplanes, Cars & 26, 21, 196 \\ 
Manufacturing & Textures ALOT, Textures DTD, Textures & 250, 47, 64 \\ 
Human Actions & MPII Human Pose, Stanford 40 Actions, 73 Sports & 29, 40, 73 \\
OCR & Omniprint-MD-6, Omniprint-MD-5-bis, Omniprint-MD-mix & 703, 706, 706 \\
\bottomrule
\end{tabularx}

\end{table}

%% file: tables/text_generator.tex
\begin{table}[h!]
\caption{Comparison between Base Model and Improved Model Responses}
\label{tabgenerator}
\centering
\resizebox{1.0\columnwidth}{!}{
\begin{tabular}{|m{5cm}|m{6.5cm}|m{6.5cm}|}
\hline
\textbf{Instruction} & \textbf{Base Model Response} & \textbf{Improved Model Response} \\
\hline
\textit{Prompt: \{"role": "system", "content": "You are an AI chatbot"\},
    \{"role": "user", "content": "Explain what AI is and its importance"\},} & 
\textbf{What is AI?} Artificial Intelligence (AI) refers to the development of computer systems that can perform tasks that would typically require human intelligence, such as:
\begin{enumerate}
  \item  \textbf{Learning}: AI systems can learn from data, experiences, and environments to improve their performance.
  \item  \textbf{Reasoning}: AI systems can reason and make decisions based on the data they have learned.
  \item  \textbf{Problem-solving}: AI systems can solve complex problems and adapt to new situations.
  \item \textbf{Perception}: AI systems can perceive and understand the world around them through sensors, cameras, and other devices.
\end{enumerate}
AI systems use various techniques, such as machine learning, deep learning, natural language processing, and computer vision, to achieve these capabilities.

\textbf{Types of AI:}
\begin{enumerate}
  \item \textbf{Narrow or Weak AI}: Designed to perform a specific task, such as facial recognition or language translation.
  \item  \textbf{General or Strong AI}: Designed to perform any intellectual task that a human can, such as reasoning, problem-solving, and learning.
  \item  \textbf{Superintelligence}: Significantly more intelligent than the best human minds, potentially leading to exponential growth in technological advancements.
\end{enumerate}
\textbf{Importance of AI:}
\begin{enumerate}
    \item \textbf{Automation}: AI can automate repetitive and mundane tasks, freeing up human resources for more strategic work.
  
\end{enumerate}

&
\textbf{What is Artificial Intelligence (AI)?} Artificial Intelligence (AI) is a branch of computer science that focuses on creating intelligent machines that can think, learn, and behave like humans. AI involves the development of algorithms, statistical models, and computer programs that enable machines to perform tasks that typically require human intelligence, such as:
\begin{enumerate}
  \item  \textbf{Reasoning and problem-solving}: AI systems can analyze data, identify patterns, and make decisions based on that data.
  \item  \textbf{Learning}: AI systems can learn from experience, adapt to new situations, and improve their performance over time.
  \item  \textbf{Perception}: AI systems can interpret and understand data from sensors, such as images, speech, and text.
  \item  \textbf{Natural Language Processing (NLP)}: AI systems can understand and generate human language, enabling them to communicate with humans.
  \end{enumerate}

\textbf{Importance of AI} AI has become increasingly important in various aspects of our lives, including:
\begin{enumerate}
  \item  \textbf{Automation}: AI can automate repetitive and mundane tasks, freeing up human time for more creative and strategic work.
  \item  \textbf{Improving decision-making}: AI can analyze large datasets, identify patterns, and provide insights that inform business decisions.
  \item  {Enhancing customer experience}: AI-powered chatbots and virtual assistants can provide 24/7 support and personalized services.
  \end{enumerate}

\\
\hline
\textit{prompt;
\{"role": "system", "content": "You work at a coffee shop"\},
    \{"role": "user", "content": "what title do you give to someone who serves the coffee to customers"\},} &
At our coffee shop, we call the person who serves the coffee to customers a 'Barista'. However, some of our
 team members are also referred to as 'Cafe Attendants' or 'Coffee Servers', depending on their specific
  role and responsibilities. But generally, the title 'Barista' is the most commonly used and well-known term
  in the industry.
&
In a coffee shop, the person who serves the coffee to customers is typically referred to as a:
Barista: This is the most common title, and it refers to a skilled person who prepares and serves coffee drinks,
as well as other beverages and food items.
Coffee Server: This title is also used to describe someone who serves coffee and other beverages to customers.
Coffee Barista or Coffee Specialist: These titles are often used in specialty coffee shops or high-end
establishments to emphasize the barista's expertise and knowledge of coffee.
In our coffee shop, I'm a barista, and I'm happy to serve you your favorite coffee drink.
\\
\hline
\end{tabular}
}

\end{table}

%% file: tables/glues_tasks_descruiption.tex
%\begin{table}
%    \centering
%        \caption{Glue benchmark tasks descriptor used in the experiment on glue datasets.}
%    \begin{tabular}{ll}
%    \toprule
%        \textbf{Task Name} & \textbf{Description} \\ 
%        \midrule
%
%        \textbf{SSTB} & Predict the similarity score between two sentences. Rate their similarity on a scale from 0 to 5, where 0 indicates no meaning overlap, 1 indicates very little overlap, and 5 indicates complete overlap in meaning.\\ 
%        \midrule
%        \textbf{MRCP} & Determine the semantic equivalence of two given sentences (Sentence 1 and Sentence 2). If the sentences are semantically equivalent, return 1. If they are not, return 0.  \\ 
%        \midrule
%        \textbf{SST2} & Determine the sentiment of a given sentence. Respond with 0 if the sentiment is negative and 1 if the sentiment is positive.   \\ 
%        \midrule
%        \textbf{COLA} & Evaluate whether the given sentence is both syntactically and semantically correct. If it is, respond with "1"; otherwise, respond with "0". \\
%       \midrule
%        \textbf{QNLI} & Evaluate whether the given response properly answers the provided question. If the response answers the question correctly, return 0; otherwise, return 1.\\ 
%        \midrule
%        \textbf{RTE} & Determine if a given hypothesis is true (entailment), false (contradiction), or undetermined (neutral) based on a provided premise.  \\
%       \bottomrule
%    \end{tabular}
%    \label{tab:glues_descriptor}
%\end{table}
%

\begin{table}[ht]
\caption{Glue benchmark tasks descriptor used in the experiment on glue datasets.}
\label{tab:glues_descriptor}
\centering
\begin{tabularx}{\textwidth}{XX}
\toprule
        \textbf{Task Name} & \textbf{Description} \\ 
        \midrule

        \textbf{SSTB} & Predict the similarity score between two sentences. Rate their similarity on a scale from 0 to 5, where 0 indicates no meaning overlap, 1 indicates very little overlap, and 5 indicates complete overlap in meaning.\\ 
        \midrule
        \textbf{MRCP} & Determine the semantic equivalence of two given sentences (Sentence 1 and Sentence 2). If the sentences are semantically equivalent, return 1. If they are not, return 0.  \\ 
        \midrule
        \textbf{SST2} & Determine the sentiment of a given sentence. Respond with 0 if the sentiment is negative and 1 if the sentiment is positive.   \\ 
        \midrule
        \textbf{COLA} & Evaluate whether the given sentence is both syntactically and semantically correct. If it is, respond with "1"; otherwise, respond with "0". \\
       \midrule
        \textbf{QNLI} & Evaluate whether the given response properly answers the provided question. If the response answers the question correctly, return 0; otherwise, return 1.\\ 
        \midrule
        \textbf{RTE} & Determine if a given hypothesis is true (entailment), false (contradiction), or undetermined (neutral) based on a provided premise.  \\
       \bottomrule
\end{tabularx}

\end{table}

%% file: tables/mixed_arch.tex
\begin{table}[t]
%\begin{wraptable}{r}{8cm}
%\vspace{-0.15in}
\caption{Evaluation on Large Datasets}
\resizebox{0.9\columnwidth}{!}{

\begin{tabular}{ccccccc}
\toprule
Datasets       & \multicolumn{2}{c}{\begin{tabular}[c]{@{}c@{}}CIFAR10\\ (ShuffleNet)\end{tabular}} & \multicolumn{2}{c}{\begin{tabular}[c]{@{}c@{}}CIFAR100\\ (ShuffleNet)\end{tabular}} & \multicolumn{2}{c}{\begin{tabular}[c]{@{}c@{}}ImageNet-1k\\ (SqueezeNet)\end{tabular}} \\ \midrule
Methods        & \multicolumn{1}{c}{Top1}                           & Top5                          & \multicolumn{1}{c}{ToP1}                           & Top5                           & \multicolumn{1}{c}{Top1}                      & Top5                                   \\ 
\midrule
Pretrained     & \multicolumn{1}{c}{92.98}                          & 99.73                         & \multicolumn{1}{c}{72.39}                          & 91.46                          & \multicolumn{1}{c}{58.178}                & \multicolumn{1}{l}{80.624}        \\ 
\midrule
Ours(sampling) & \multicolumn{1}{c}{\textbf{93.14 $\pm$ 0.25}}          & \textbf{99.76$\pm$ 0.22}           & \multicolumn{1}{c}{\textbf{72.60 $\pm$ 0.15}}           & \textbf{91.29 $\pm$ 0.13}           & \multicolumn{1}{c}{\textbf{58.257 $\pm$ 1.022}}        & \textbf{81.01$\pm$ 1.251}                     \\
\bottomrule
\end{tabular}

}
\label{table:mixed}
%\vskip -0.10in
\end{table}
%\end{wraptable}

%% file: tables/zerishot_res18_in_dist.tex
\begin{table}[t]
\centering
\caption{\textbf{Zero-Shot Transfer Learning} This Table represent results of zero-shot evaluation against the pretrained model on Resnet18 full model architecture.}
\resizebox{0.6\columnwidth}{!}{
\begin{tabular}{lccc}
\toprule
Model & MNIST & CIFAR-10 & CIFAR-100\\
\midrule
Pretrained & 99.61& 94.56  &75.86\\
D2NWG(ours) & 99.62 $\pm$ 0.07 & 94.57 $\pm$ 0.00 & 75.83 $\pm$ 0.02\\ 
\bottomrule
\end{tabular}

}
\label{tab:res18}
\end{table}

%% file: tables/phi3.tex
\begin{table}[h]
\centering
\caption{Generating weights for the Microsoft Phi-3 language model output head.}
    \resizebox{0.8\textwidth}{!}{
\begin{tabular}{lcccc}
\toprule
Methods & ARC Challenge (25-shots) & ARC Easy (25-shots) & HellaSwag (10-shots) & Winogrande (5-shots) \\
\midrule
Pretrained & 87.16 $\pm$ 0.00 & 63.23 $\pm$ 0.01 & 73.65 $\pm$ 0.01 & 76.64 $\pm$ 0.01 \\
\ourmethod     & 87.36 $\pm$ 0.01 & 63.74 $\pm$ 0.01 & 73.65 $\pm$ 0.00 & 76.72 $\pm$ 0.01 \\
\bottomrule
\end{tabular}
}
\label{phi3}
\end{table}

%% file: tables/gtp2.tex
\begin{table}[t]
\caption{Performance evaluation on unseen open llms leaderboard v2 benchmark base on full gpt2-164M small. These results are produced by Huggingface after submission to open LLM leaderdoards. $\uparrow $ indicate performance improvement while $\downarrow$ indicate a performance decrease}
\label{gpt2}
\centering
\resizebox{\textwidth}{!}{
\renewcommand{\arraystretch}{1.0}
\renewcommand{\tabcolsep}{4pt}
\begin{tabular}{lccccccccc}
\toprule
    % \textbf{Method} & \textbf{ifeval (0 shot)} & \textbf{Bbh (3 shots)} & \textbf{Gpqa (0 shots)} & \textbf{MATH-hard (4)} & \textbf{Musr (0 shot)} & \textbf{MMLU-Pro (5 shots)} & \textbf{Avg} \\ 
    \textbf{Method} & \textbf{ifeval (0)} & \textbf{Bbh (3)} & \textbf{Gpqa (0)} & \textbf{MATH-hard (4)} & \textbf{Musr (0)} & \textbf{MMLU-Pro (5)} & \textbf{Avg}&Base Model& Fine-tuned \\ 
\midrule

openai-community-gpt2  & 17.8  &  2.83 & \textbf{1.12} & 0.3 & \textbf{13.91}  & 1.84 & 6.3& na & Yes\\ 
\midrule
\ourmethod      & \textbf{19.16}($\uparrow $\textcolor{blue}{1.36}) &\textbf{2.85}($\uparrow $\textcolor{blue}{0.02}) & 1.01($\downarrow $\textcolor{blue}{0.11})  & \textbf{0.38}($\uparrow $\textcolor{blue}{0.08}) & 12.68($\downarrow $\textcolor{blue}{1.23}) & \textbf{1.68}($\downarrow $\textcolor{blue}{0.16}) & 6.29($\downarrow $\textcolor{blue}{0.01}) &openai-community-gpt2 & No \\ 

% \ourmethod      & \textbf{19.89}($\uparrow $\textcolor{blue}{1.36}) &\textbf{2.39}($\downarrow $\textcolor{blue}{0.02}) & 1.12($\downarrow $\textcolor{blue}{0.11})  & \textbf{0.12}($\downarrow $\textcolor{blue}{0.08}) & 13.24($\downarrow $\textcolor{blue}{1.23}) & \textbf{1.50}($\downarrow $\textcolor{blue}{0.16}) & 6.29($\downarrow $\textcolor{blue}{0.01}) &openai-community-gpt2 & No \\ 

\bottomrule
\end{tabular}}

\end{table}

%% file: figures/hyperzooplots.tex
 \begin{figure*}[t!]
      \centering
    \begin{subfigure}[b]{0.24\columnwidth}
          \centering
          \includegraphics[width=\textwidth]{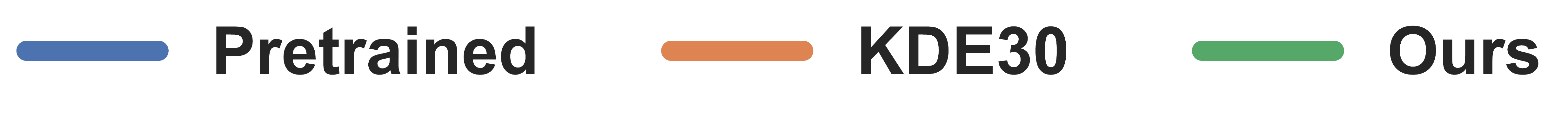}
      \end{subfigure}
      
      \begin{subfigure}[b]{0.24\columnwidth}
          \centering
          \includegraphics[width=\textwidth]{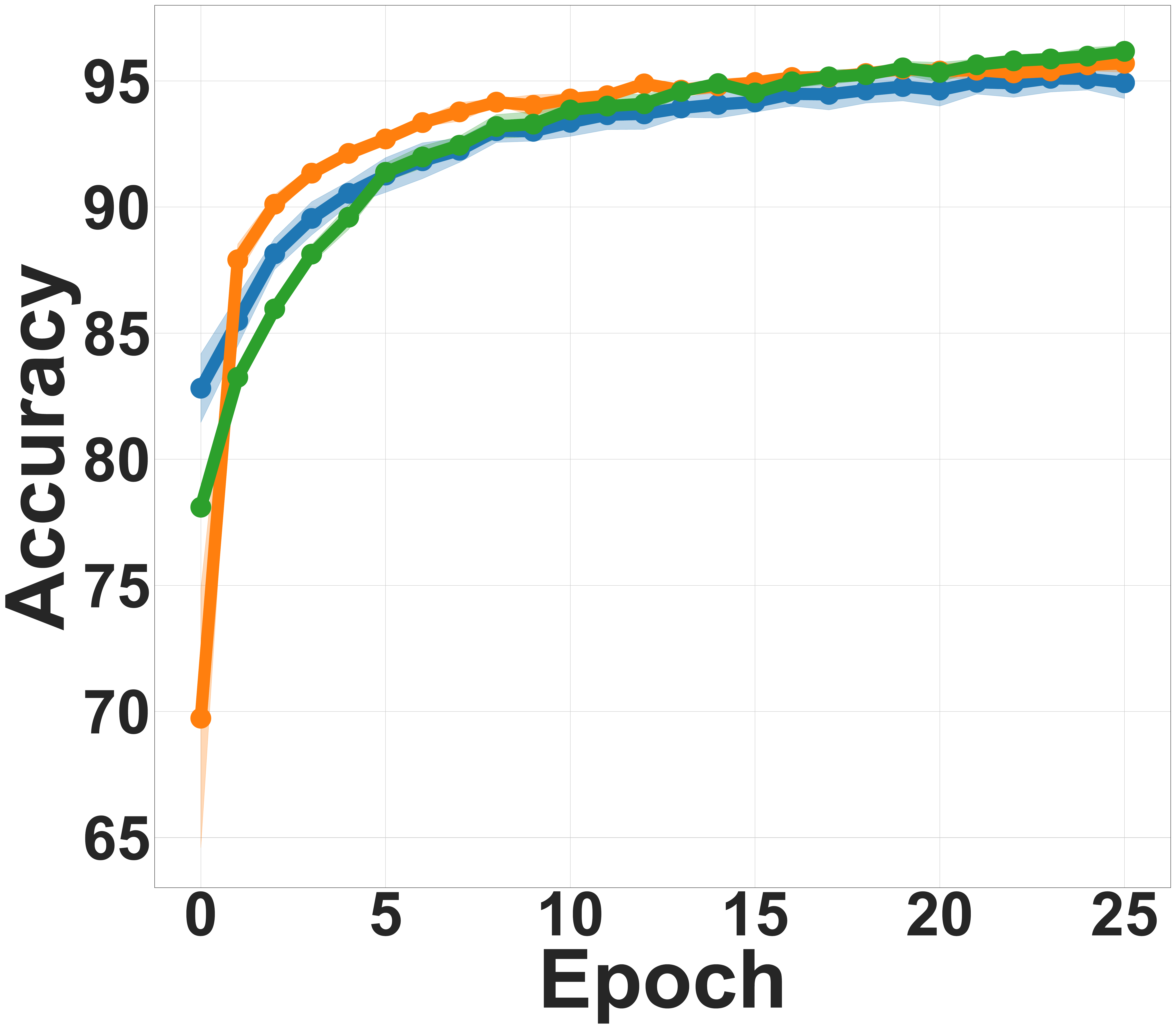}
          \caption{MNIST}
          \label{figmnist}
      \end{subfigure}
      \begin{subfigure}[b]{0.24\columnwidth}
          \centering
          \includegraphics[width=\textwidth]{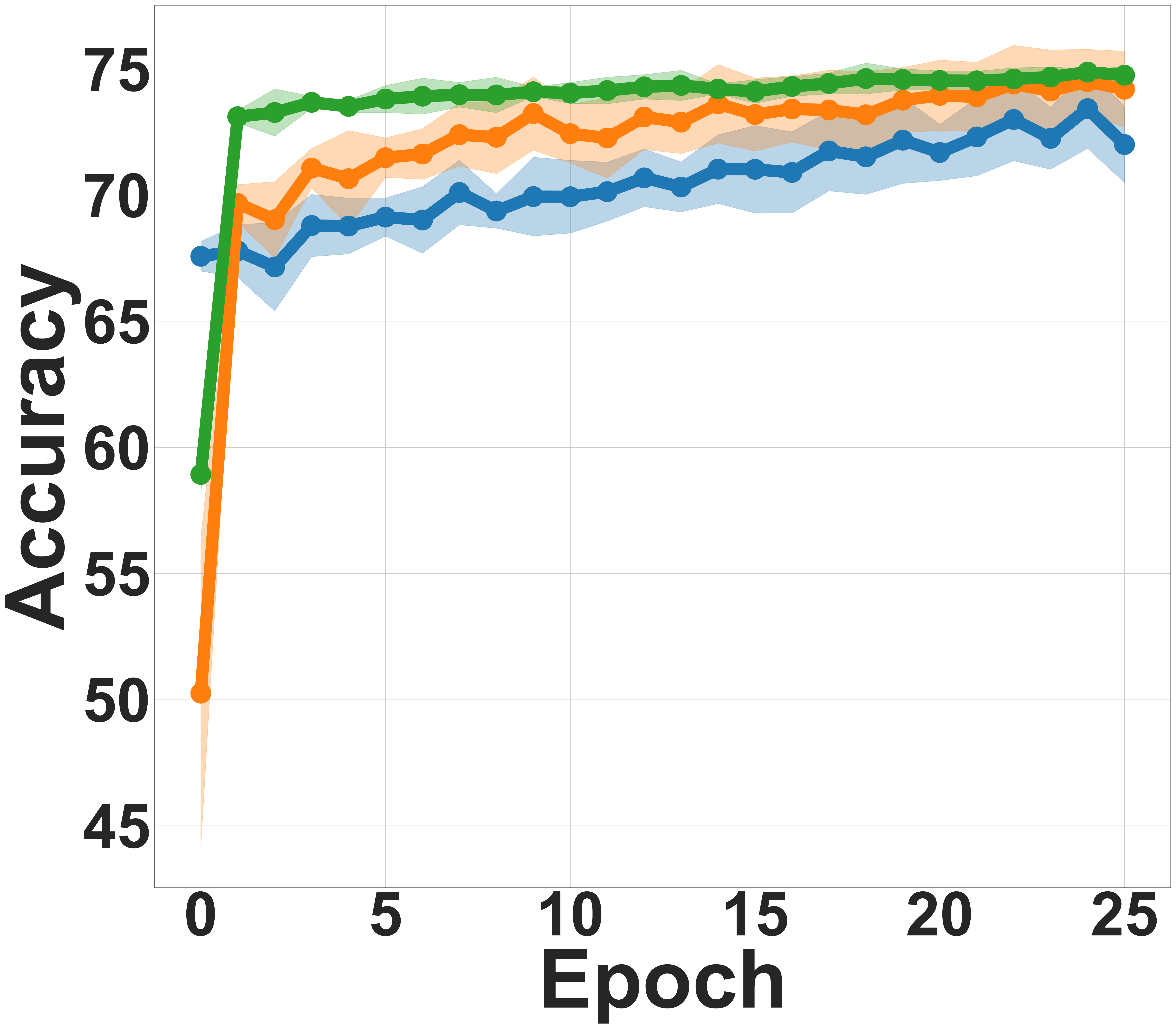}
          \caption{SVHN}
          \label{figsvhn}
      \end{subfigure}
      \begin{subfigure}[b]{0.24\columnwidth}
         \centering
             \includegraphics[width=\textwidth]{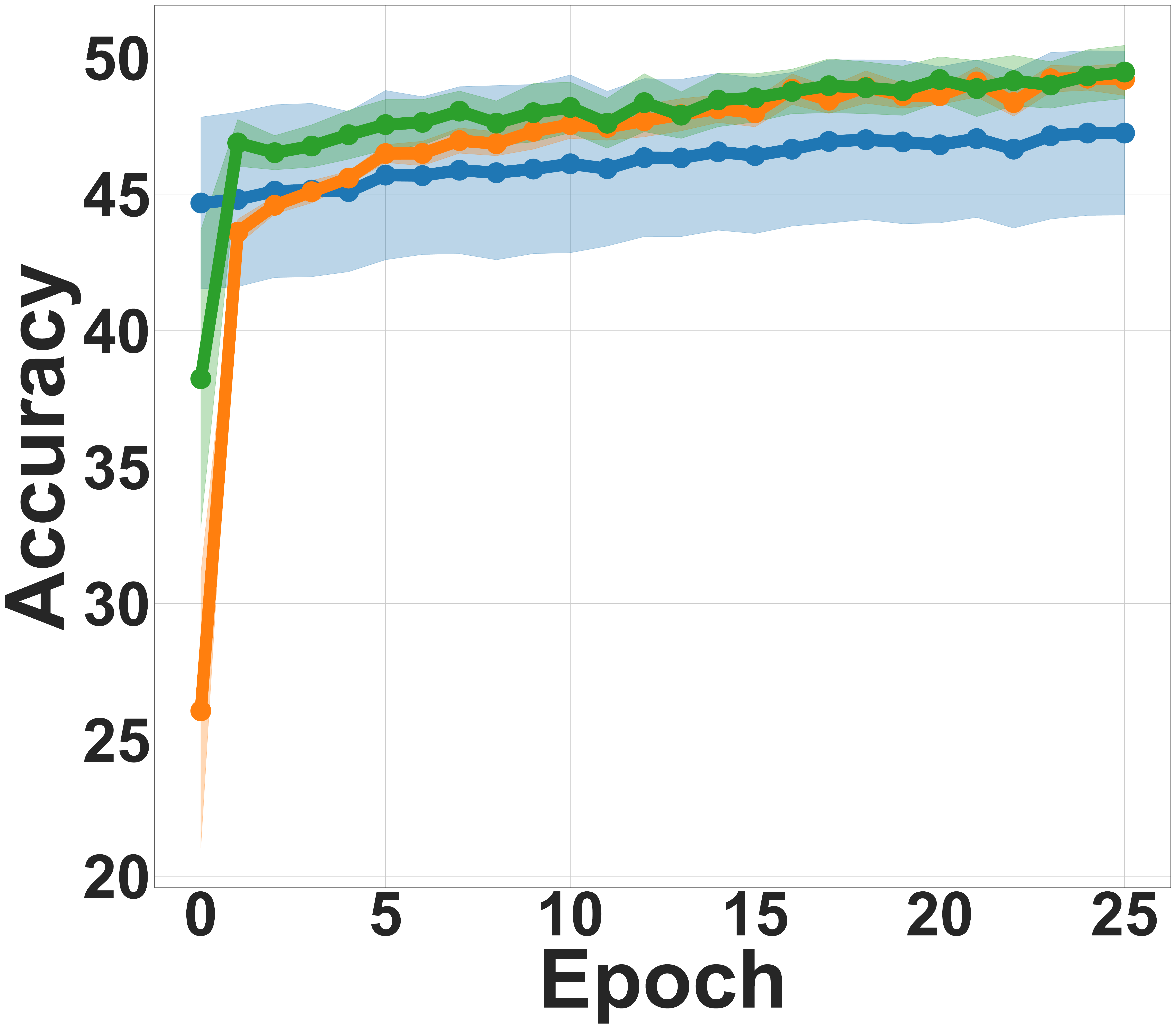}
         \caption{CIFAR-10}
         \label{figcifar}
      \end{subfigure}
      % \hfill
      \begin{subfigure}[b]{0.24\columnwidth}
          \centering
          \includegraphics[width=\textwidth]{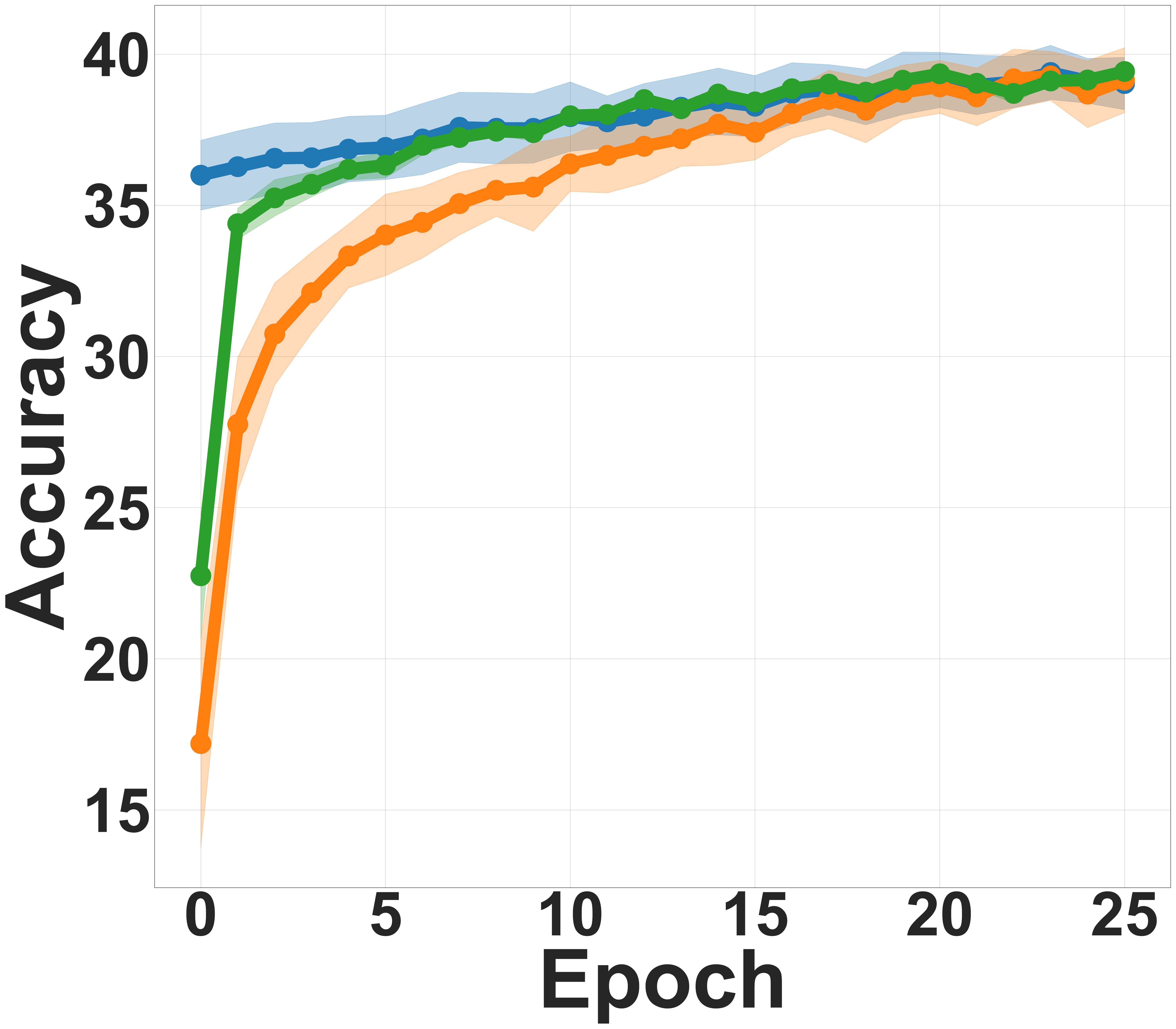}
          \caption{STL-10}
          \label{figstl}
      \end{subfigure}
         \caption{\textbf{Convergence Plots on Finetuning Generated Weigths:}
         Weights generated by the competing methods are finetuned for 25 epochs on the training set. We utilize the modelzoos of ~\citet{schurholtModelZoosDataset2022}.         }
         \label{neuripsevol}
        \vspace{-0.15in}
 \end{figure*}

%% file: tables/dataset_transfer.tex
\begin{table}[t]
% \begin{wraptable}{r}{0.5\columnwidth}
% \vspace{-0.15in}
\caption{\textbf{No Fine-tuning Initialization on Unseen Datasets} We transfer from one dataset, or combinations of datasets, to unseen datasets at test time.}\label{wtab}
% \vspace{-0.25in}
\centering
\resizebox{0.6\columnwidth}{!}{
\begin{tabular}{lcccc}\\
\toprule  
Source & Target& Accuracy & Methods\\
\midrule
MNIST& SVHN&  13.25& \multirow{4}{*}{S\textsubscript{\texttt{KDE30}}} \\
SVHN& MNIST &  29.30& \\
CIFAR-10 & STL-10 &  15.20&  \\
STL-10& CIFAR-10 &  15.40& \\
\midrule
\multicolumn{4}{c}{Sampling from Combined Weights Distribution}\\
\midrule
MNIST+CIFAR-10& SVHN & 18.80 &\multirow{4}{*}{Ours}\\
MNist+CIFAR-10& STL-10 & 16.21 &\\
SVHN + STL-10& MNIST & 36.64 & \\
SVHN + STL-10& CIFAR-10 &18.00 &\\
\bottomrule
\end{tabular}
}
\vskip -0.10in
\end{table} 
% \end{wraptable}

%% file: tables/appendixtable2.tex
\begin{table*}[t]
\caption{Performance evaluation at initialization without fine-tuning. For the baseline we use weights of SVHN for MNIST and vice versa similarly for CIFAR-10 and STL-10}\label{wtab22}
\begin{center}
\resizebox{0.8\textwidth}{!}{
\begin{tabular}{ccccc}\\
\toprule  
Datasets & MNIST & SVHN & CIFAR10 & STL10 \\
\midrule
Random &10.23$\pm$0.56 & 12.21$\pm$3.76 &9.98$\pm$1.47 & 9.56$\pm$1.02\\  
Pretrained models&  82.82$\pm$ 1.38& \textbf{67.57$\pm$ 0.59}& \textbf{44.68$\pm$ 3.15} & \textbf{35.99$\pm$ 1.15}\\
S\textsubscript{kde30}\citet{schurholtHyperRepresentationsGenerativeModels2022} &  69.73$\pm$ 5.12& 50.25$\pm$ 6.12 & 26.06$\pm$ 3.01& 17.20$\pm$ 3.43\\
\midrule
seen (D2NWG) &83.92$\pm$1.92 & 61.81 $\pm$ 3.13& 43.08$\pm$0.55 & 31.45$\pm$0.35 \\ 
seen(D2NWG)(with Pred) &84.85$\pm$0.83 & 66.03 $\pm$ 1.36& 43.89$\pm$0.15 & 34.29$\pm$0.13 \\ 
\midrule
S\textsubscript{kde30}\citet{schurholtHyperRepresentationsGenerativeModels2022}(cross) &  29.30$\pm$ 3.46& 13.25$\pm$ 1.12 & 15.40$\pm$ 0.51& 15.20$\pm$1.24 \\
% \midrule
not seen(D2NWG) &36.64$\pm$4.69 & 18.80$\pm$0.58& 18.00$\pm$0.22& 16.21$\pm$0.52\\ 
not seen(D2NWG)(with Pred) &30.15$\pm$5.09 & 15.76$\pm$1.43& 17.10$\pm$1.12& 15.37$\pm$0.52\\ 
\bottomrule
\end{tabular}
}
\end{center}
\vskip -0.10in
\end{table*} 

%% file: tables/appendixtable4.tex
\begin{table*}[h!]
\caption{In-distribution performance comparison of different image dataset encoding schemes on model zoo dataset}\label{tab:dataencoder}
\centering
\resizebox{0.8\linewidth}{!}{
\begin{tabular}{ccccc}\\
\toprule  
Datasets & MNIST & SVHN & CIFAR10 & STL10 \\
\midrule
Pretrained models&  82.82$\pm$ 1.38& \textbf{67.57$\pm$ 0.59}& \textbf{44.68$\pm$ 3.15} & \textbf{35.99$\pm$ 1.15}\\
S\textsubscript{kde30}\citet{schurholtHyperRepresentationsGenerativeModels2022} &  69.73$\pm$ 5.12& 50.25$\pm$ 6.12 & 26.06$\pm$ 3.01& 17.20$\pm$ 3.43\\
\midrule
MLP\_Encoder &67.04$\pm$17.73 & 35.65 $\pm$ 13.03& 17.41$\pm$3.02 & 20.36$\pm$7.38 \\ 
Set\_transf(pret) &78.21$\pm$1.76 & 60.90 $\pm$ 1.08& 28.68$\pm$1.84 & 34.75$\pm$00.38 \\ 
seen (D2NWG) &83.92$\pm$1.92 & 61.81 $\pm$ 3.13& 43.08$\pm$0.55 & 31.45$\pm$0.35 \\ 
seen(D2NWG)(with Pred) &84.85$\pm$0.83 & 66.03 $\pm$ 1.36& 43.89$\pm$0.15 & 34.29$\pm$0.13 \\  
\bottomrule
\end{tabular}
}
\end{table*}

%% file: tables/appendixtable3.tex
\begin{table*}[t]
    \caption{Performance of the datasets conditional sampling on 10 unseen real-world datasets. We report the averaged accuracy on ten unseen test datasets over 3 different runs fine-tuned for 50 epochs. pret(imnet): pretrained on imagenet1k}
    \begin{center}
    \resizebox{1.0\textwidth}{!}{% <------ Don't forget this %
    \begin{tabular}{l|ccc|ccc|c}
    \toprule
       \multirow{2}{*}{Datasets} & \multicolumn{3}{c|}{No-fine-tuning} & \multicolumn{3}{c}{50 epochs Fine-Tuning}&  \multirow{2}{*}{\# of classes} \\
        % \hline
         & Random init.& pret(imnet) & D2NWG(ours) & Random init.& pret(imnet) & D2NWG(ours)& \\
        %\hline
        \midrule
        % \midrule
        Gemstones & 1.13 $\pm$ 0.52 & 0.62 $\pm$ 0.00& \textbf{1.86 $\pm$ 0.25} & 70.59$\pm$ 0.91 & 67.49$\pm$ 0.43 & \textbf{76.06 $\pm$ 0.88} & 87\\
         %\hline
        Dog Breeds &0.55 $\pm$ 0.22 & 0.69 $\pm$ 0.00& \textbf{1.87 $\pm$ 0.39} & 80.78$\pm$ 0.28 & 78.13$\pm$ 0.49 & \textbf{80.88 $\pm$ 0.88} & 133\\
         %\hline
        Dessert& 21.03 $\pm$ 2.44 & 12.50 $\pm$ 0.00  & \textbf{99.40 $\pm$0.02} & 95.83$\pm$0.34 & 94.64$\pm$ 0.00 & \textbf{99.40 $\pm$ 0.02} & 5 \\
         %\hline
        Colorectal Histology& 11.77 $\pm$2.88 & 11.00 $\pm$ 0.00 & \textbf{18.12 $\pm$ 0.25} & 90.34 $\pm$ 0.33 & 89.75$\pm$ 0.19 & \textbf{93.65 $\pm$ 0.10} & 8 \\
         %\hline
       Drawing&10.86 $\pm$ 1.22 & 11.00 $\pm$ 0.00 & \textbf{11.87 $\pm$0.93} & \textbf{90.20 $\pm$ 0.16} & 90.00$\pm$ 0.16 & 89.00 $\pm$ 0.16 & 10 \\
        % \hline
        Alien vs Predator&51.48 $\pm$2.09 & 28.88 $\pm$ 0.00 & \textbf{78.15 $\pm$0.52} & 98.52$\pm$ 0.52 & \textbf{98.89$\pm$ 1.42} & 97.77 $\pm$ 0.00 & 2 \\
        % \hline
        COVID-19&20.13 $\pm$18.66 & 46.53 $\pm$ 0.00 & \textbf{47.22 $\pm$0.00} & 93.86$\pm$0.16 & 93.40$\pm$ 0.49 & \textbf{94.56 $\pm$ 0.71} & 3\\
        % \hline
        honey-bee-pollen&49.54 $\pm$1.30 & 50.00 $\pm$ 0.00 &\textbf{56.94 $\pm$4.53} & 93.05 $\pm$ 0.00 & 88.89$\pm$ 0.00 & \textbf{93.55  $\pm$ 4.53} & 2 \\
        % \hline
        Speed Limit Signs& 30.55 $\pm$2.27 & 25.00 $\pm$ 0.00 &\textbf{31.48 $\pm$10.23} & 83.33$\pm$ 0.00 &  86.11$\pm$ 0.00 & \textbf{90.74 $\pm$ 1.31} & 4 \\
        % \hline
        Japanese Characters&0.03$\pm$0.00 & 0.08 $\pm$ 0.00 & \textbf{0.50$\pm$0.22} & 53.17 $\pm$ 0.15 &  \textbf{62.33 $\pm$ 0.16} & 62.16 $\pm$0.47 0.45 & 1566\\
         \bottomrule
    \end{tabular}
    }
    \end{center}
    \label{tab22}
    % \vskip -0.10in
\end{table*}

%% file: figures/vit_bar.tex
\begin{figure*}
  \begin{minipage}[b]{.65\linewidth}
    \centering
\resizebox{1.0\columnwidth}{!}{
\begin{tabular}{lcccc}
\toprule
Model & MNIST& SVHN & CIFAR-10 & STL-10\\
\midrule
Pretrained & 99.42 $\pm$ 0.05 & 94.62 $\pm$ 0.18 & 93.51 $\pm$ 0.16  &94.01 $\pm$ 0.10\\
Linear\_prob & 96.88 $\pm$ 0.45 & 57.23 $\pm$ 0.28 & 82.85 $\pm$ 0.25  &95.63 $\pm$ 1.23\\
\midrule
D2NWG(ful) &  \textbf{99.55} $\pm$ \textbf{0.02} & \textbf{95.13} $\pm$ \textbf{0.10} & \textbf{94.23} $\pm$ \textbf{0.27} &94.02 $\pm$ 0.10\\
D2NWG(rob) &  97.56 $\pm$ 0.26 & 57.41 $\pm$ 0.17 & 83.64 $\pm$ 0.47  &\textbf{95.74} $\pm $ \textbf{0.74}\\
\midrule
Cross datasets transfer learning\\
\midrule
OFA (Pretrained)\cite{cai2020once}& 13.34& 8.90&  13.34&  8.90\\
D2NWG(full) &  \textbf{66.82} $\pm$ \textbf{0.65} & \textbf{35.20} $\pm$ \textbf{0.65} & \textbf{36.70} $\pm$ \textbf{0.18}  & \textbf{51.50} $\pm$ \textbf{0.37}\\
D2NWG(prob) &  42.86 $\pm$ 0.62 & 20.974 $\pm$ 0.78 & 26.56 $\pm$ 1.22  &47.33 $\pm$ 0.32\\
\bottomrule
\end{tabular}	

   }
    \captionof{table}{\small MobileNet Weight Generation.}%\textbf{Full-model weights generation} This Table represents results of zero-shot evaluation against the pretrained model on mobilenetv3 full model architecture weights sampling versus classifier layer weights sampling}
    \label{tab:mbvv3}
    
  \end{minipage}\hfill
  \begin{minipage}[b]{.35\linewidth}
    \centering
    \includegraphics[width=0.9\linewidth]{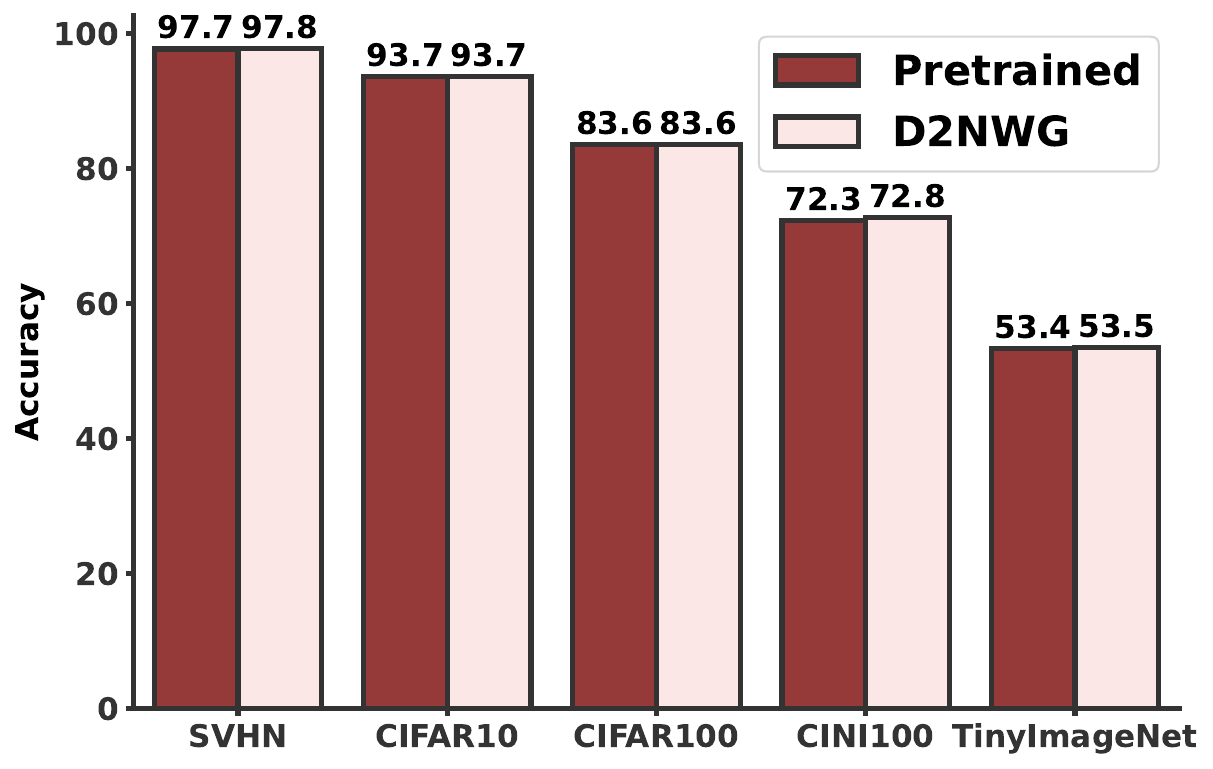}
    \captionof{figure}{\small Experiment with ViT}% architectures~\citep{Gani_2022_BMVC}.}% \caption{Figure caption}
    \label{barv1}
  \end{minipage}
\end{figure*}

%% file: figures/similarity.tex
 \begin{figure}[t!]
      \centering
    \begin{subfigure}[b]{0.4\columnwidth}
          \centering
          \includegraphics[width=\textwidth]{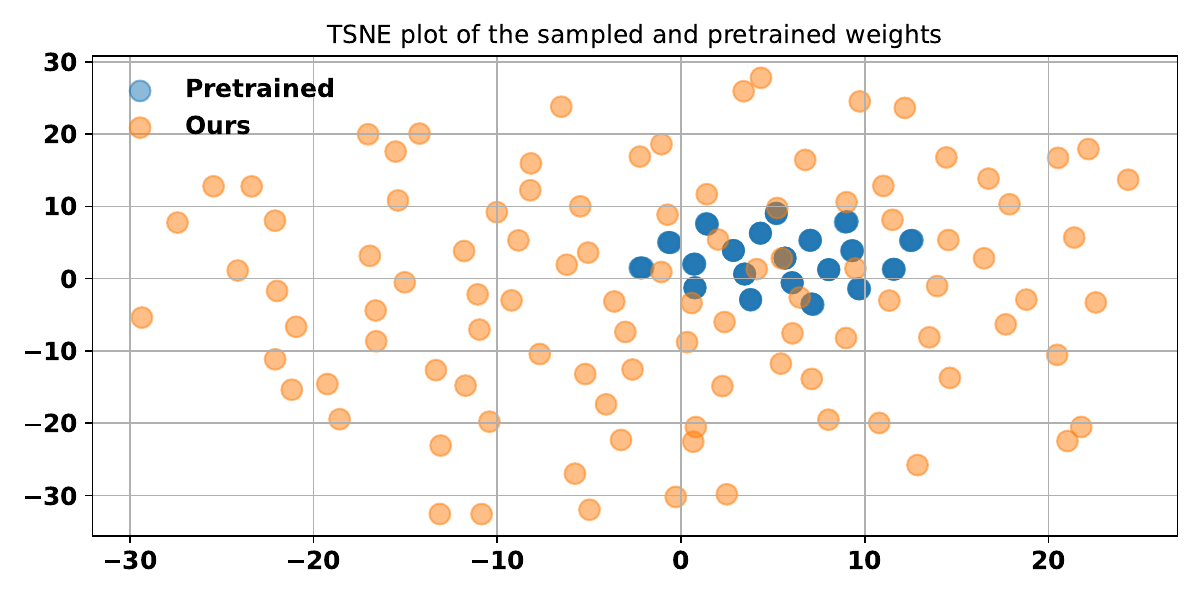}
          \caption{TSNE}
          \label{tsne}
      \end{subfigure}
      
      \begin{subfigure}[b]{0.4\columnwidth}
          \centering
          \includegraphics[width=\textwidth]{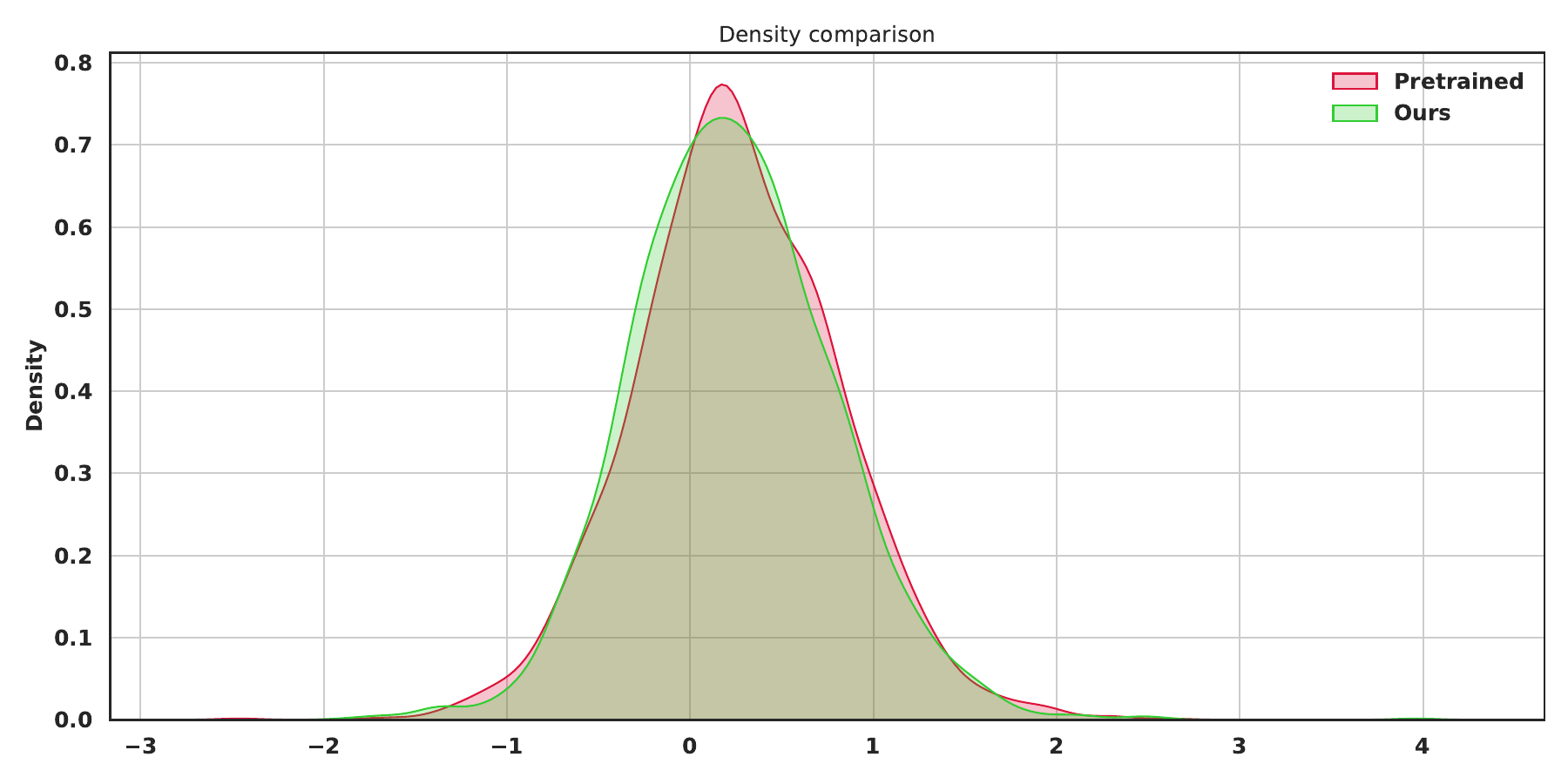}
          \caption{Density}
          \label{kde}
      \end{subfigure}

      \begin{subfigure}[b]{0.4\columnwidth}
         \centering
             \includegraphics[width=\textwidth]{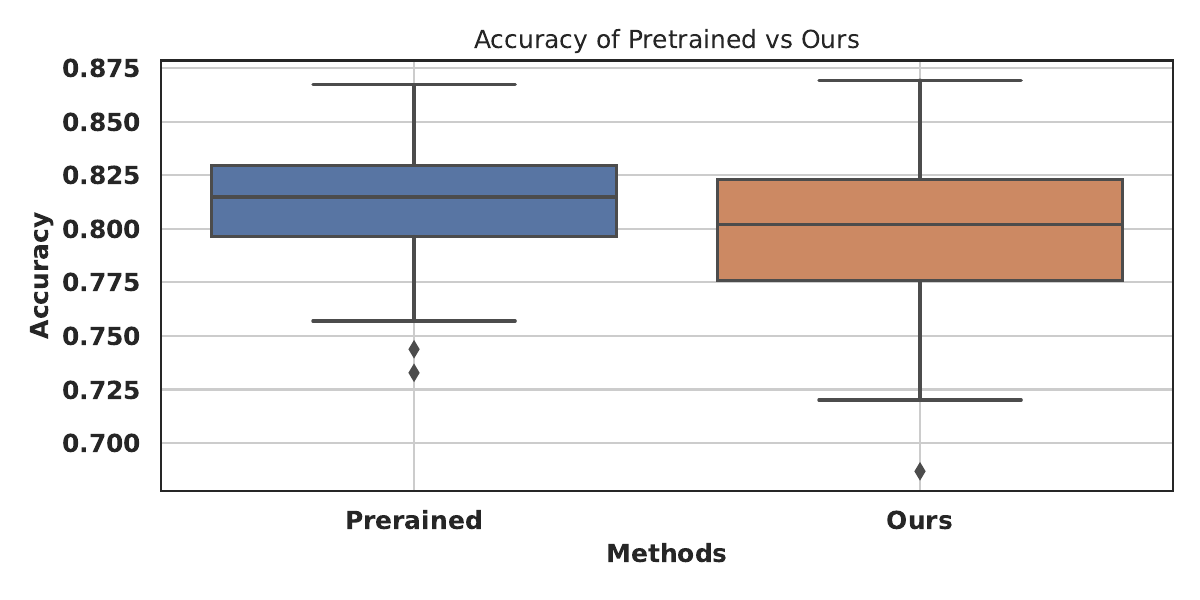}
         \caption{Accuracy}
         \label{acc}
      \end{subfigure}

         \caption{\textbf{Analysis of relationship between the pretrained weights and the sampled weights for MNIST dataset}}
         \label{sim}
        \vspace{-0.1in}
 \end{figure}